\documentclass[11pt]{article}

\usepackage[final]{acl_spherelab}
\aclfirstpagefooter{~\raisebox{-0.38ex}{\includegraphics[height=.34cm]{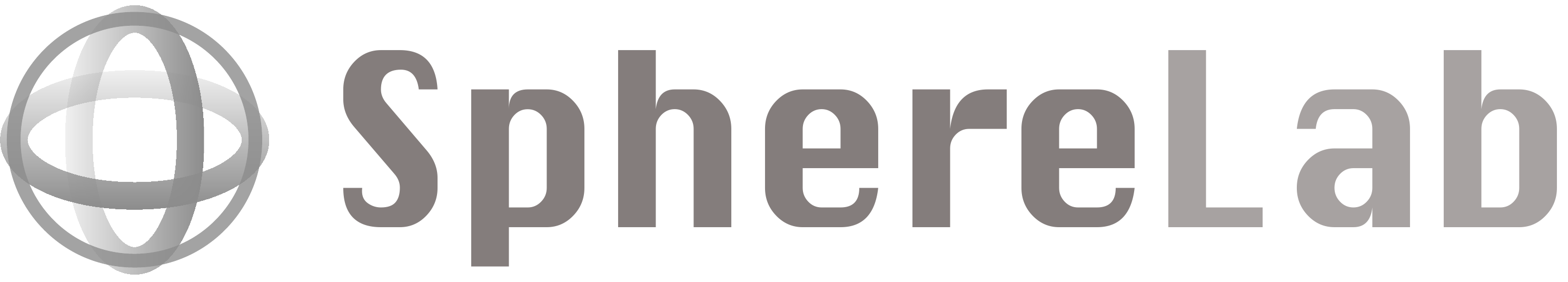}}~~~\fontsize{8.75pt}{\baselineskip}\selectfont Technical Report}

\usepackage{times}
\usepackage{latexsym}
\usepackage[T1]{fontenc}
\usepackage[utf8]{inputenc}
\usepackage{microtype}
\usepackage{inconsolata}
\usepackage{amsfonts}
\usepackage{graphicx}
\usepackage{booktabs}
\usepackage{wrapfig}
\usepackage{amsmath}
\usepackage{amsthm}
\usepackage{amssymb}
\usepackage{supertabular}
\usepackage{algpseudocode}
\usepackage[linesnumbered,lined,boxed,commentsnumbered,ruled,longend]{algorithm2e}
\theoremstyle{plain}

\theoremstyle{definition}

\theoremstyle{remark}

\usepackage{enumitem}
\usepackage{subcaption}
\usepackage{textcomp}
\usepackage{enumerate}

\definecolor{deepred}{HTML}{940000}
\hypersetup{linkcolor=deepred}
\hypersetup{urlcolor  = [rgb]{0.4,0.15,0.95}}
\hypersetup{citecolor=[rgb]{0.4,0.15,0.95}}
\usepackage{url}

\usepackage[capitalize,noabbrev]{cleveref}

\usepackage{epigraph}
\usepackage{tikz}
\usetikzlibrary{shadows}
\usepackage[align=center,shadow=true,shadowsize=4.5pt,nobreak=true,framemethod=tikz,skipabove=9.5pt,skipbelow=9pt,innertopmargin=6pt,innerbottommargin=6pt,innerleftmargin=5pt,innerrightmargin=5pt,leftmargin=1.5pt,rightmargin=1.5pt]{mdframed}
\usepackage{tcolorbox}
\usepackage{multirow}
\usepackage{makecell}
\usepackage{xspace}

\newlength\savewidth\newcommand\shline{\noalign{\global\savewidth\arrayrulewidth
  \global\arrayrulewidth 1pt}\hline\noalign{\global\arrayrulewidth\savewidth}}

\SetKwInput{KwInput}{Input}
\SetKwInput{KwOutput}{Output}

\tcbset{
  myexample/.style={
    colback=white,
    colframe=black!60,
    colbacktitle=black!80,
    coltitle=white,
    fonttitle=\bfseries,
    arc=4mm,
    boxrule=1pt,
    left=3mm, right=3mm, top=1.5mm, bottom=1.5mm,
    title={#1}
  }
}

\definecolor{caseblue}{HTML}{5C6FA8}

\tcbset{
  casestudy/.style={
    colback=white,
    colframe=black!60,
    colbacktitle=caseblue,    %
    coltitle=white,
    fonttitle=\bfseries,
    arc=4mm,
    boxrule=1pt,
    left=3mm, right=3mm, top=1.5mm, bottom=1.5mm,
    title={#1}
  }
}

\newcommand{\ie}{\emph{i.e.}}
\newcommand{\eg}{\emph{e.g.}}

\usepackage{pifont}
\usepackage{tocloft}
\usepackage[toc,page,header]{appendix}
\usepackage{adjustbox}
\usepackage{minitoc}

\renewcommand \thepart{}
\renewcommand \partname{}

\title{PEFT-Arena: Understanding Parameter-Efficient Finetuning\\from a Stability-Plasticity Perspective}

\definecolor{skyblue}{RGB}{135, 206, 235}
\definecolor{palegreen}{RGB}{152, 251, 152}
\definecolor{revblue}{RGB}{0, 82, 155}
\definecolor{revred}{RGB}{160, 32, 48}

\author{
Yangyi Huang$^{1,\dagger}$ \quad Ruotian Peng$^{2,\dagger}$ \quad Zeju Qiu$^3$ \quad Jiale Kang$^1$ \quad Yandong Wen$^2$\\[0.5mm]
\textbf{Bernhard Sch\"olkopf}$^3$ \quad \textbf{Weiyang Liu}$^{1,3,*}$ \\[1mm]
$^1$The Chinese University of Hong Kong~~~$^2$Westlake University~~~$^3$MPI for Intelligent Systems \\[0.5mm]
$^\dagger$Equal contribution~~~~$^*$Corresponding author~~~~\href{https://spherelab.ai/PEFT-Arena}{\tt SphereLab.ai/PEFT-Arena} \\
}

\long\def\paperabstract{%

Parameter-efficient finetuning (PEFT) has become the standard approach for adapting large language models, yet evaluations largely emphasize downstream accuracy while overlooking the retention of pretrained capabilities. We argue that PEFT should be assessed through the stability-plasticity dilemma: the trade-off between target-task adaptation and resistance to forgetting. We introduce PEFT-Arena, a benchmark that jointly measures downstream performance and general capability retention. Across methods, we find distinct stability-plasticity profiles; under comparable parameter budgets, orthogonal finetuning achieves the most favorable Pareto frontier.
To explain these differences, we analyze PEFT updates from two geometric perspectives. In weight space, spectral analysis reveals how parameterizations interact with the pretrained singular-value structure. In activation space, retention metrics show whether finetuning preserves or distorts general-capability representations, with forgetting linked to non-isometric representation distortion. 
Finally, an analysis shows that final SFT checkpoints often overshoot a better target-retention operating point. Inspired by this, we present case studies of a post-hoc improvement with path-wise rewinding.

}

\begin{document}

\maketitle

\doparttoc 
\faketableofcontents

\begin{abstract}
\paperabstract

\end{abstract}

\begin{figure*}[t]
\centering
  \vspace{7mm}
  \setlength{\abovecaptionskip}{5pt}
  \setlength{\belowcaptionskip}{2pt}
  \includegraphics[width=\textwidth]{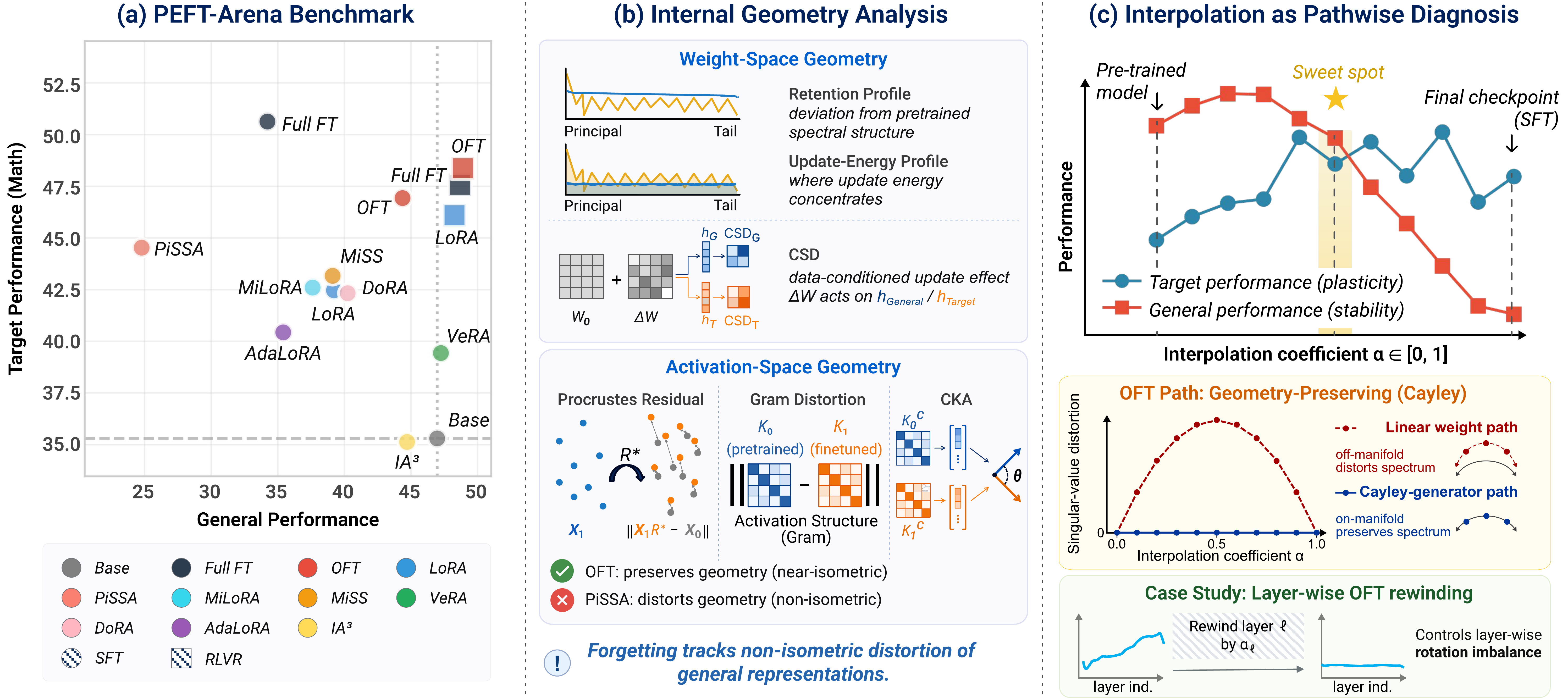}
  \caption{PEFT-Arena is designed to comprehensively evaluate the trade-off between downstream task adaptation and pretrained knowledge retention in LLM post-training with PEFT methods. 
  (a) External stability--plasticity trade-offs across PEFT methods. 
  (b) Internal geometry analysis from weight-space and activation-space views. 
  (c) Interpolation reveals SFT overshoot and motivates pathwise rewinding along method-specific update paths.}
  \label{fig:teaser}
  \vspace{-1.5mm}
\end{figure*}

\vspace{-.5mm}
\section{Introduction}
\vspace{-1mm}

Parameter-efficient finetuning (PEFT) has become essential for adapting large foundation models to downstream tasks. By updating only a small subset of parameters, PEFT enables practical, low-cost deployment across diverse domains. But how should we determine whether a PEFT method is truly effective? Current practice often reduces this question to a single metric (\ie, downstream task performance), while overlooking what the finetuned model may lose in the process. However, this single-metric paradigm can be misleading. A method that substantially improves mathematical reasoning while silently degrading instruction following, factual recall, and general reasoning by a comparable margin has not truly adapted the model; rather, it has broken it. Although numerous PEFT methods have demonstrated strong effectiveness on downstream tasks, the extent to which they preserve pretrained capabilities after adaptation remains largely unclear. Therefore, the real question is not only ``how much did the model learn?'', but also ``how much did it learn relative to how much it forgot?'' This is precisely characterized by the stability-plasticity dilemma~\cite{mermillod2013stability}: the tension between acquiring new capabilities (plasticity) and preserving existing ones (stability). Guided by this dilemma, we are interested in the question below:

\vspace{-1.5mm}
\begin{mdframed}
    [leftmargin=0.2em, rightmargin=0.2em] \fontsize{10pt}{\baselineskip} \itshape
    \centering
    Which PEFT method yields the most favorable stability-plasticity trade-off?
    \vspace{-.5mm}
\end{mdframed}
\vspace{-3mm}

To this end, we introduce \textbf{PEFT-Arena}, a benchmark that jointly measures target-domain performance (plasticity) and general capability retention (stability) across two challenging reasoning domains, mathematics and medicine. With PEFT-Arena, we find that neither target performance nor general performance alone is sufficient for PEFT evaluation. All methods exhibit stability-plasticity trade-offs, but different parameterizations exhibit distinct trade-off patterns. In particular, orthogonal finetuning (OFT) often lies on a strong frontier, suggesting that the geometry of the update plays an important role in preserving general capabilities.

Benchmark results reveal the trade-offs induced by PEFT, but shed no light on how PEFT reshapes the model internally. It motivates another question:

\vspace{-1.5mm}
\begin{mdframed}
    [leftmargin=0.2em, rightmargin=0.2em] \fontsize{10pt}{\baselineskip} \itshape
    \centering
    Which internal mechanisms are associated with the stability-plasticity trade-off?
    \vspace{-.5mm}
\end{mdframed}
\vspace{-3mm}

We approach this question from two complementary views. In \emph{weight space}, we examine how PEFT updates interact with the pretrained spectral geometry of weight matrices. This view highlights the inductive bias of each parameterization: additive low-rank methods, spectral-initialization variants, and orthogonal transformations reshape the pretrained basis in distinct ways. In \emph{activation space}, we examine the representations induced by the finetuned model on the same evaluation examples. The key issue is not simply whether activations move, but whether finetuning preserves the relative structure among examples that the pretrained model represented coherently. We measure this non-isometric distortion using Procrustes residual, pairwise Gram distortion, and linear CKA. This view links forgetting to representation-geometry damage. We find that OFT better preserves the structure of general representation than other PEFT methods.

Finally, we use interpolation as a pathwise diagnostic of finetuning dynamics:

\vspace{-1.5mm}
\begin{mdframed}
    [leftmargin=0.2em, rightmargin=0.2em] \fontsize{10pt}{\baselineskip} \itshape
    \centering
    Can interpolation reveal where PEFT loses stability along the adaptation path?
    \vspace{-.5mm}
\end{mdframed}
\vspace{-2.7mm}

Weight interpolation between the base and finetuned model exposes a common SFT overshoot phenomenon, \ie, the final checkpoint often moves beyond the best target-retention operating point. We use interpolation as a diagnostic tool to study the stability-plasticity trade-off. Moreover, interpolation must respect each PEFT method's natural update geometry. For additive methods, the natural path scales the additive update $\Delta W$; for OFT, the natural path scales the skew-symmetric Cayley generator rather than linearly interpolating dense weights. Within this view, layer-wise OFT rewinding serves as a practical example of post-hoc control for imbalanced update strength. Our contributions are listed below:

\begin{itemize}[leftmargin=*,nosep]
    \setlength{\itemsep}{0.37em}

    \item \textbf{A multi-faceted PEFT benchmark}. We evaluate PEFT methods based on both target-domain gains and general capability preservation.

    \item \textbf{Findings on PEFT trade-off patterns}. We show that PEFT methods exhibit distinct stability-plasticity behavior, and that OFT often provides a strong frontier under the same parameter budget.

    \item \textbf{Internal analysis from weight and activation geometry}. We connect external forgetting to internal changes through two empirical views: spectral profiles of weight updates and non-isometric distortion of activation geometry.

    \item \textbf{Interpolation as pathwise diagnosis}. We use interpolation to diagnose SFT overshoot and emphasize parameterization-aware interpolation paths, with OFT's Cayley path and layer-wise rewinding illustrating geometry-aware control.
\end{itemize}

\vspace{-0.3mm}
\section{The PEFT-Arena Benchmark}
\vspace{-0.65mm}

\subsection{Experimental Setup}
\vspace{-0.1mm}
\label{sec:experiments:setup}
All PEFT methods are evaluated along two axes: (i) target-domain performance and (ii) general ability retention. We use target-domain performance as a proxy for \emph{plasticity}, and general-task performance as a proxy for \emph{stability}. We conduct experiments on two target domains, mathematics and medicine, under two post-training settings: supervised finetuning (SFT) and reinforcement learning with verifiable rewards (RLVR). Unless otherwise stated, we report average accuracy (\%) for each domain.

\textbf{Target-domain benchmarks and evaluation.}
  \emph{(i) Math}: We evaluate on a combined set of Math-500~\cite{lightman2023lets}, AMC23, and AIME24. 
  \emph{(ii) Medicine}: We evaluate on a collection of medical reasoning and knowledge benchmarks, including MedMCQA~\cite{pmlr-v174-pal22a}, MedQA (USMLE)~\cite{app11146421}, PubMedQA~\cite{jin2019pubmedqa}, MMLU-Pro~\cite{NEURIPS2024_ad236edc}, GPQA (Medical)~\cite{rein2024gpqa}, Lancet, NEJM \& MedBullets problems~\cite{chen-etal-2025-benchmarking}, and MedXpertQA~\cite{zuo_medxpertqa_2025} following a dedicate dataset survey~\cite{huang_m1_nodate}. 

\textbf{General ability retention.}
To measure general ability preservation after adaptation, we evaluate on IFEval~\cite{zhou2023instructionfollowingevaluationlargelanguage}, NQ~\cite{kwiatkowski_natural_2019}, BBH~\cite{suzgun2022challenging}, covering instruction following, natural language understanding, general knowledge and general reasoning. We use the average score ($\%$) across these tasks as our General score to assess model forgetting after finetuning. We follow an OpenCompass-style evaluation configuration with context length 1024, temperature $T=0.0$, and one sample per query.

\textbf{Base models and adaptation methods.}
We use Qwen2.5-7B~\cite{qwen2024qwen2} and Llama3.2-3B-Instruct~\cite{dubey2024llama} as pretrained LLMs to cover different scales and base/instruction-tuned settings.
We compare full finetuning (Full FT) against a representative set of PEFT baselines (see Appendix~\ref{app:related-work} for related work details).
\emph{(i) Additive PEFT (LoRA family)}:
We include LoRA~\cite{hu2022lora} and representative variants spanning rank allocation, parameterization and initialization: AdaLoRA~\cite{zhang2023adaptive}, DoRA~\cite{liu2024dora}, MiSS~\cite{kang2025missrevisitingtradeofflora}, VeRA~\cite{kopiczko2024vera}, PiSSA~\cite{meng_pissa_2024}, and MiLoRA~\cite{wang2025miloraharnessingminorsingular}.
We also include KeepLoRA~\cite{luo2026keeplora}, an anti-forgetting LoRA variant that constrains updates away from the principal subspace.
\emph{(ii) Multiplicative PEFT}:
We include orthogonal finetuning (OFT)~\cite{qiu2023controlling,liu_parameter-efficient_2024,qiu_orthogonal_2025}, which constrains updates to structured orthogonal transformations with adjustable sparsity.
\emph{(iii) Activation-based PEFT}:
We include IA$^3$~\cite{liu2022few}, a lightweight method that adapts models via learned activation scaling.

To fairly compare PEFT method under a similar parameter budget, \autoref{tab:main-results} includes budget-matched SFT slices rather than a single configuration per method. On Qwen, the roughly 20M trainable-parameter group compares OFT-b32 (17.55M) with LoRA/PiSSA/MiLoRA/KeepLoRA-r8 (20.19M) and DoRA-r8 (21.58M), while the roughly 40M group compares OFT-b64 (35.68M) with LoRA-r16 (40.37M), with the Llama columns reporting the corresponding backbone-specific counts.

\textbf{Training and optimization details.}
We conduct SFT in both target domains, using 50k filtered samples from OpenR1-Math-330k~\cite{huggingface2025openr1} for math and 23k samples from m23k~\cite{huang_m1_nodate} for medical. 
We also include RLVR results with GRPO~\cite{shao2024deepseekmathpushinglimitsmathematical} on a representative subset of methods for comparison. Full details are provided in Appendix~\ref{app:implementation}.

\definecolor{maingain}{RGB}{0,102,51}
\definecolor{mainloss}{RGB}{160,32,48}
\newcommand{\ppdelta}[1]{%
  \begingroup
  \edef\ppdeltatmp{#1}%
  \ifnum\pdfstrcmp{\ppdeltatmp}{+0.00}=0
    \textcolor{black!65}{(#1)}%
  \else
    \expandafter\ppdeltaaux\ppdeltatmp\relax
  \fi
  \endgroup
}
\def\ppdeltaaux#1#2\relax{%
  \if#1+%
    \textcolor{maingain}{(#1#2)}%
  \else
    \textcolor{mainloss}{(#1#2)}%
  \fi
}

\begin{table*}[t]
  \centering
  \footnotesize
  \vspace{-1mm}
  \setlength{\abovecaptionskip}{7pt}
  \setlength{\belowcaptionskip}{5pt}
  \setlength{\tabcolsep}{2pt}
  \renewcommand{\arraystretch}{1.29}
  \resizebox{\textwidth}{!}{%
  \begin{tabular}{ll|ccccc|ccccc}
    \multirow{2}{*}{Method} & \multirow{2}{*}{Settings} & \multirow{2}{*}{\shortstack{Tr. Param\\(Qwen)}} & \multicolumn{4}{c|}{\textbf{Qwen2.5-7B-base}} & \multirow{2}{*}{
    \shortstack{Tr. Param\\(Llama)}
    } & \multicolumn{4}{c}{\textbf{Llama3.2-3B-Instruct}} \\
    & & & Math Target (\%) & Math General (\%) & Med Target (\%) & Med General (\%) & & Math Target (\%) & Math General (\%) & Med Target (\%) & Med General (\%) \\[.5mm]
    \shline
    \multicolumn{12}{c}
    {\textbf{\textit{Supervised FineTuning (SFT)}}} \\[.5mm]

    Base & -- & 7.61B & 35.30 \ppdelta{+0.00} & 46.97 \ppdelta{+0.00} & 46.36 \ppdelta{+0.00} & 46.97 \ppdelta{+0.00} & 3.21B & 27.80 \ppdelta{+0.00} & 53.03 \ppdelta{+0.00} & 41.44 \ppdelta{+0.00} & 53.03 \ppdelta{+0.00} \\
    Full FT & -- & 7.61B & 50.63 \ppdelta{+15.33} & 34.22 \ppdelta{-12.74} & 53.63 \ppdelta{+7.27} & 34.41 \ppdelta{-12.56} & 3.21B & 33.90 \ppdelta{+6.27} & 39.83	\ppdelta{-13.20} & 44.26 \ppdelta{+2.82} & 26.03 \ppdelta{-27.00} \\[.5mm]

    \hline
    OFT & block 16 & 8.49M & 42.33 \ppdelta{+7.03} & 42.58 \ppdelta{-4.39} & 46.17 \ppdelta{-0.19} & 45.09 \ppdelta{-1.88} & 7.08M & 29.43 \ppdelta{+1.80} & 41.08 \ppdelta{-11.95} & 39.22 \ppdelta{-2.22} & 40.97 \ppdelta{-12.06} \\
    OFT & block 32 & 17.55M & 46.93 \ppdelta{+11.63} & 44.37 \ppdelta{-2.60} & 48.63 \ppdelta{+2.27} & 42.40 \ppdelta{-4.57} & 11.55M & 30.60 \ppdelta{+2.97} & 40.73 \ppdelta{-12.30} & 39.50 \ppdelta{-1.94} & 40.50 \ppdelta{-12.53} \\
    OFT & block 64 & 35.68M & 46.23 \ppdelta{+10.93} & 35.97 \ppdelta{-11.00} & 49.47 \ppdelta{+3.11} & 39.11 \ppdelta{-7.86} & 24.97M & 29.30 \ppdelta{+1.67} & 39.75 \ppdelta{-13.28} & 40.77 \ppdelta{-0.67} & 37.70 \ppdelta{-15.33} \\
    OFT & block 128 & 71.92M & 47.77 \ppdelta{+12.47} & 36.98 \ppdelta{-9.99} & 52.40	\ppdelta{+6.04} & 36.88	\ppdelta{-10.08} & 47.34M & 32.23 \ppdelta{+4.60} & 36.26	\ppdelta{-16.76} & 42.17 \ppdelta{+0.73} & 34.26 \ppdelta{-18.77} \\[.5mm]

    \hline
    LoRA & r4a8 & 10.09M & 42.33	\ppdelta{+7.03}&				41.66	\ppdelta{-5.31} & 47.12	\ppdelta{+0.76}	 & 36.42	\ppdelta{-10.55}& 6.9M & 24.30	\ppdelta{-3.33} &	35.79	\ppdelta{-17.23} &		36.92	\ppdelta{-4.52} &	31.84	\ppdelta{-21.19} \\
    LoRA & r8a16 & 20.19M & 42.47 \ppdelta{+7.17} & 39.22 \ppdelta{-7.75} & 47.91 \ppdelta{+1.55} & 36.06 \ppdelta{-10.91} & 12.16M & 24.07 \ppdelta{-3.56} & 36.57 \ppdelta{-16.46} & 38.34 \ppdelta{-3.10} & 27.99 \ppdelta{-25.04} \\
    LoRA & r16a32 & 40.37M &44.87	\ppdelta{+9.57} &	34.91	\ppdelta{-12.06} &		47.86	\ppdelta{+1.51}	 & 34.86	\ppdelta{-12.11}& 24.31M & 24.97	\ppdelta{-2.66} & 37.55	\ppdelta{-15.48} &39.21	\ppdelta{-2.23} & 	29.18	\ppdelta{-23.85}\\
    LoRA & r32a64 & 80.74M & 45.37 \ppdelta{+10.07} & 38.21 \ppdelta{-8.76} & 49.48 \ppdelta{+3.12} & 35.50 \ppdelta{-11.47} & 48.63M  & 25.90	\ppdelta{-1.73} & 37.20	\ppdelta{-15.83}&				39.33	\ppdelta{-2.11} &		30.69	\ppdelta{-22.34} \\[.5mm]
    \hline
    AdaLoRA & r8a16 & 30.28M & 40.43 \ppdelta{+5.13} & 35.41 \ppdelta{-11.56} & 45.22 \ppdelta{-1.13} & 37.34 \ppdelta{-9.63} & 18.24M & 20.83 \ppdelta{-6.80} & 34.53 \ppdelta{-18.49} & 37.11 \ppdelta{-4.33} & 36.29 \ppdelta{-16.74} \\
    PiSSA & r8a16 & 20.19M & 44.53 \ppdelta{+9.23} & 24.78 \ppdelta{-22.19} & 26.16 \ppdelta{-20.19} & 18.05 \ppdelta{-28.92} & 12.16M & 0.67	\ppdelta{-26.96} & 9.74 \ppdelta{-43.28} & 21.17	\ppdelta{-20.27}	& 12.92	\ppdelta{-40.11} \\
    MiLoRA & r8a16 & 20.19M  & 42.60	\ppdelta{+7.30} &	37.62 \ppdelta{-9.35} & 46.83	\ppdelta{+0.48}	& 35.88	\ppdelta{-11.09} & 12.16M & 23.60 \ppdelta{-4.03} &	35.59 \ppdelta{-17.44} & 37.64	\ppdelta{-3.81} &	29.23	\ppdelta{-23.80} \\
    KeepLoRA & r8 & 20.19M & 40.53 \ppdelta{+5.23} & 43.75 \ppdelta{-3.22} & 45.60 \ppdelta{-0.76} & 47.09 \ppdelta{+0.12} & 12.16M & 15.20 \ppdelta{-12.43} & 40.74 \ppdelta{-12.29} & 41.26 \ppdelta{-0.18} & 39.52 \ppdelta{-13.51} \\
    MiSS & r8 & 11.12M & 43.17 \ppdelta{+7.87} & 39.12 \ppdelta{-7.85} & 48.75 \ppdelta{+2.40} & 34.43 \ppdelta{-12.54} & 6.19M & 23.37 \ppdelta{-4.26} & 33.93 \ppdelta{-19.09} & 40.16 \ppdelta{-1.28} & 31.71 \ppdelta{-21.32} \\
    MiSS & r64 & 89.00M & 46.93 \ppdelta{+11.63} & 32.77 \ppdelta{-14.20} & 51.90 \ppdelta{+5.54} & 32.72 \ppdelta{-14.25} & 49.55M & 28.63 \ppdelta{+1.00} & 34.96 \ppdelta{-18.06} & 41.96 \ppdelta{+0.52} & 22.78 \ppdelta{-30.25} \\
    VeRA & r256 & 1.44M & 39.43 \ppdelta{+4.13} & 47.25 \ppdelta{+0.38} & 28.50 \ppdelta{-17.85} & 47.01 \ppdelta{+0.04} & 0.82M & 28.80	\ppdelta{+1.17} &	46.79	\ppdelta{-6.23}	&	40.68	\ppdelta{-0.76}	& 48.94	\ppdelta{-4.09} \\
    DoRA & r8a16 & 21.58M & 42.33 \ppdelta{+7.03} & 40.25 \ppdelta{-6.72} & 48.04 \ppdelta{+1.69} & 36.06 \ppdelta{-10.91} & 12.93M & 23.83 \ppdelta{-3.80} & 35.65 \ppdelta{-17.37} & 38.25 \ppdelta{-3.19} & 27.53 \ppdelta{-25.50} \\
    IA$^3$ & -- & 1.82M & 35.13 \ppdelta{-0.17} & 44.71 \ppdelta{-2.26} & 30.08 \ppdelta{-16.28} & 48.25 \ppdelta{+1.28} & 0.92M & 29.70 \ppdelta{+2.07} & 45.72	\ppdelta{-7.30}	&	39.13	\ppdelta{-2.31}	& 45.67	\ppdelta{-7.36} \\[.5mm]

    \shline
    \multicolumn{12}{c}{
    \textbf{\textit{RLVR with Group Relative Policy Optimization (GRPO)}}} \\[.5mm]
    Full FT & -- & 7.61B & 47.57	\ppdelta{+12.27} &	48.68	\ppdelta{+1.71} &		46.24	\ppdelta{-0.11} &	43.22	\ppdelta{-3.75} & 3.21B  & 29.80	\ppdelta{+2.17} &	52.20	\ppdelta{-0.82} &		45.88	\ppdelta{+4.44} &	51.81	\ppdelta{-0.83} \\
    OFT & block 32 & 17.55M & 47.90	\ppdelta{+12.60} &	48.90	\ppdelta{+1.93}	& 46.79	\ppdelta{+0.44} &	47.24	\ppdelta{+0.27} & 11.55M & 29.97	\ppdelta{+2.34}	 &50.04	\ppdelta{-2.98}		 & 44.99	\ppdelta{+3.55} &	52.31	\ppdelta{-2.99} \\
    LoRA & r8a16 & 20.19M & 46.93	\ppdelta{+11.63} &	48.27	\ppdelta{+1.30}  & 47.08	\ppdelta{+0.73} &	42.80	\ppdelta{-4.17} &  12.16M &28.83	\ppdelta{+1.20} &	52.17	\ppdelta{-0.85} &		44.97	\ppdelta{+3.53} & 53.53	\ppdelta{-0.86} \\
  \end{tabular}%
  }
  \vspace{-1mm}
  \caption{Main benchmark results. For each domain, average task accuracy is reported in \% (higher is better). We also report the absolute change relative to the corresponding base model in parentheses. Trainable-parameter columns show the comparable parameter budget for different methods.}
  \label{tab:main-results}
  \vspace{-2.5mm}
\end{table*}

\subsection{Main Results and Discussions}

We report benchmark results along two axes: target-domain performance (plasticity) and general ability retention (stability), under both SFT and RLVR settings. The complete results are given in \autoref{tab:main-results}. In the following, we summarize the key empirical findings.
Unless otherwise specified, all changes relative to the corresponding base model are absolute differences in percentage points.

\textbf{SFT improves target performance at the expense of general ability}.
In \autoref{tab:main-results}, Full FT gives the largest target gains but it also incurs the most severe forgetting. On Qwen2.5-7B, Full FT increases the math target accuracy from 35.30 to 50.63 and the medical target accuracy from 46.36 to 53.63, while the general performance drops from 46.97 to 34.22 for math and drops from 46.97 to 34.41 for medicine. On Llama3.2-3B-Instruct, the general performance falls from 53.03 to 26.03 for medicine. The results suggest that target-only reporting systematically overestimates post-training quality.

\textbf{Under SFT, methods show distinct trade-off patterns, with OFT on the best frontier}.
Within the additive low-rank family, LoRA, MiSS, DoRA, and AdaLoRA generally improve target performance but tend to incur non-trivial forgetting, with larger adaptation capacity usually pushing further toward plasticity. For example, on Qwen math, LoRA-r8 improves target by 7.17 with a 7.75 general drop, while MiSS-r64 reaches 11.63 target gain with a 14.20 general drop. SVD-guided variants (MiLoRA and especially PiSSA), which rely on initialization or subspace selection, are less stable in this benchmark: PiSSA-r8 improves Qwen math target by 9.23 but drops Qwen math general by 22.19 and Qwen medical target by 20.19. The anti-forgetting LoRA variant, KeepLoRA, partially improves knowledge retention: on Qwen it raises math general from 39.22 (LoRA-r8) to 43.75 and even preserves medical general ability at 47.09, but its target adaptation is much weaker and it does not dominate the frontier across settings, especially on Llama. This suggests that retention-oriented subspace constraints alone do not guarantee the strongest overall trade-off. Outside LoRA-style methods, IA$^3$ (activation scaling) and VeRA (shared frozen projection matrices with a small number of trainable scaling vectors) are both highly parameter-efficient and relatively conservative: VeRA preserves Qwen general ability best (math/medical general: +0.38/+0.04) but sacrifices medical target performance (-17.85), while IA$^3$ shows a similar low-plasticity profile. In contrast, OFT's spectrum-preserving multiplicative parameterization gives the best balance between adaptation and retention: OFT-b32 improves Qwen math target by 11.63 with only a 2.60 drop on math general, forming the strongest stability-plasticity frontier among PEFT baselines.
This frontier is not only a comparison across methods but also across comparable trainable-parameter budgets: in the roughly 20M Qwen group, OFT-b32 is compared against LoRA/PiSSA/MiLoRA/KeepLoRA-r8 and DoRA-r8, and in the roughly 40M group, OFT-b64 is compared against LoRA-r16.

\vspace{-0.2mm}
\textbf{RLVR generally enables stable adaptation while causing less forgetting}.
Compared with SFT, RLVR with GRPO exhibits a qualitatively different regime. On Qwen math adaptation, Full FT, OFT, and LoRA improve target by 12.27, 12.60, and 11.63, while their math-general scores also increase by 1.71, 1.93, and 1.30. OFT remains slightly above Full FT on target performance (47.90 vs. 47.57) with far fewer trainable parameters (17.55M vs. 7.61B), while LoRA reaches 46.93 with 20.19M trainable parameters. This behavior is consistent with on-policy optimization, where updates are anchored to the model's own trajectories; under this regime, structured PEFT parameterizations can better capture the RL objective efficiently without large functional drift.

\vspace{-0.2mm}
\textbf{Longer GRPO training reveals a related high-$k$ degradation pattern}. \label{sec:rl-longer-training}
From \autoref{tab:main-rl-longer-passk}, we observe that longer GRPO training reveals a related pathwise degradation pattern under high-$k$ evaluation. Pass@1 target performance remains relatively stable, but high-$k$ sampling can degrade after extended optimization, with Full FT and LoRA showing larger pass@64 drops than OFT. This resembles SFT over-adaptation from a different evaluation angle. We revisit this pathwise view in \autoref{sec:interpolation}, where interpolation diagnoses SFT overshoot; Appendix~\ref{app:rlvr-highk-interp} further suggests that interpolation can also partially recover longer-RLVR high-$k$ degradation. Beyond the main General axis, Appendix~\ref{app:expanded-general} reports expanded validation on HumanEval, HellaSwag, WinoGrande, MMLU(avg), ARC, and GSM8K. These additional benchmarks are consistent with the General axis in Table~\ref{tab:main-results} and broaden our coverage of general capabilities.

\begin{table*}[t]
  \centering
  \footnotesize
  \setlength{\abovecaptionskip}{3pt}
  \setlength{\belowcaptionskip}{2pt}
  \setlength{\tabcolsep}{4.75pt}
  \renewcommand{\arraystretch}{1.18}
  \begin{tabular}{lcccccccc}
    \multirow{2}{*}{\textbf{Method}} & \multicolumn{2}{c}{\textbf{AIME24}} & \multicolumn{2}{c}{\textbf{AMC23}} & \multicolumn{2}{c}{\textbf{MATH500}} & \multicolumn{2}{c}{\textbf{AVG}} \\
    & pass@1 & pass@64 & pass@1 & pass@64 & pass@1 & pass@64 & pass@1 & pass@64 \\
    \shline
    Full FT & 13.9 / 15.1 & \textbf{50.0} / 40.0 & 52.7 / 55.1 & \textbf{92.5} / 87.5 & \textbf{76.1} / \textbf{76.6} & 93.2 / 93.4 & 47.57 / 48.93 & \textbf{78.57} / 73.63 \\
    LoRA & 11.4 / 14.8 & 43.3 / 36.7 & \textbf{54.6} / \textbf{57.4} & \textbf{92.5} / \textbf{92.5} & 74.8 / 75.5 & 94.4 / 92.2 & 46.93 / \textbf{49.23} & 76.73 / 73.80 \\
    OFT  & \textbf{16.5} / \textbf{15.4} & 46.7 / \textbf{43.3} & 52.8 / 51.9 & 90.0 / 90.0 & 74.4 / 76.2 & \textbf{95.0} / \textbf{94.0} & \textbf{47.90} / 47.83 & 77.23 / \textbf{75.77} \\
  \end{tabular}
  \caption{Longer-RLVR high-$k$ evaluation. Each entry denotes the result at Step 200 (left) and Step 500 (right).}
  \label{tab:main-rl-longer-passk}
  \vspace{-1mm}
\end{table*}

\vspace{-1.25mm}
\begin{mdframed}
    [leftmargin=0.2em, rightmargin=0.2em] \fontsize{10pt}{\baselineskip} \itshape

\textbf{Takeaway}. In SFT, PEFT method exhibits drastically different stability-plasticity trade-offs. SFT is less resistant to forgetting than RLVR. Among all methods, OFT often lies on the favorable frontier, highlighting the value of spectrum preservation.
    
\vspace{-.5mm}
\end{mdframed}
\vspace{-2.25mm}

The benchmark shows \emph{what} trade-offs occur, but it does not explain \emph{how} different PEFT parameterizations change the model internally. Next, we propose to analyze PEFT updates through weight-space and activation-space geometry, with a focus on how these changes affect general capabilities.

\section{Understanding PEFT Updates through Internal Geometry}
\label{sec:spectral_analysis}
\label{sec:internal_analysis}

PEFT-Arena exposes external stability-plasticity trade-offs, but benchmark scores alone do not explain how different parameterizations preserve or disrupt general capabilities. We therefore analyze PEFT updates from two complementary internal views. The \emph{weight-space} view examines how updates interact with the spectral geometry of pretrained parameters. The \emph{activation-space} view measures how much finetuning distorts the pairwise structural similarity of representations induced by general-evaluation data, which provides a direct view of capability retention.

\begin{figure*}[t!]
  \centering
  \setlength{\abovecaptionskip}{2.5pt}
  \setlength{\belowcaptionskip}{5pt}
  \includegraphics[width=\textwidth]{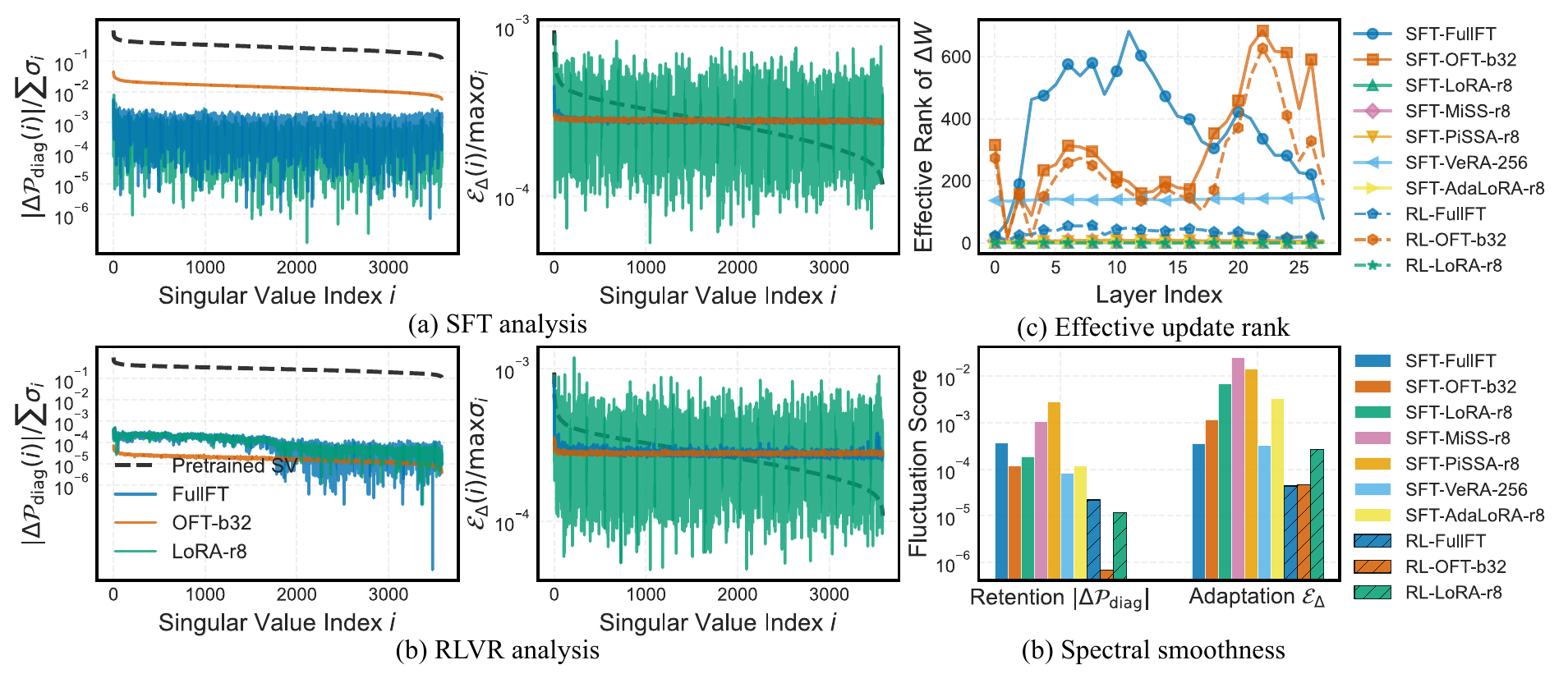}
  \caption{Weight-space spectral retention-adaptation profiling. (a \& b) Distributions of projected spectrum changes (retention and adaptation profiles). (c) Effective rank of weight updates on all \texttt{down\_proj} layers. (d) Fluctuation scores that quantify spectral smoothness of retention and adaptation profiles.}
  \label{fig:spectral_analysis}
  \vspace{-3mm}
\end{figure*}

\subsection{Weight-Space Geometry}
\label{sec:weight-geometry}

Inspired by prior work~\cite{biderman2024lora,zhu2025path,mukherjee2025reinforcement,martin2021implicit}, we start by analyzing PEFT updates in the pretrained spectral basis.
We use two descriptive measures of weight-space geometry: a \emph{retention} profile, which measures how much the pretrained singular structure is preserved, and an \emph{adaptation} profile, which measures where update energy is injected.
These profiles characterize the update geometry induced by different PEFT parameterizations; in the next subsection, we complement them with activation-space diagnostics that measure whether the resulting representations preserve the structure of general-evaluation data.

Let the pretrained weight be decomposed as $W_{0}=U\Sigma V^{\top}$, where $u_i$ and $v_i$ are the $i$-th left and right singular vectors. We study a finetuned weight $W^{*}$ from two complementary views.

\textbf{Retention profile: diagonal projection on the pretrained basis.}
We measure how much $W^{*}$ preserves the pretrained singular alignment via
\begin{equation}
\begin{aligned}
  P_{\text{diag}}(i) &= u_{i}^{\top}W^{*}v_{i}, \\
  \left | \Delta P_{\text{diag}}(i) \right |
  &=\left |u_{i}^{\top}(W^{*}-W_{0})v_{i}\right |.
\end{aligned}
\end{equation}
The quantity $|\Delta P_{\text{diag}}(i)|$ measures component-wise deviation from the pretrained singular structure. Large or irregular changes indicate stronger perturbation of principal directions that may support pretrained general capabilities.

\textbf{Adaptation profile: update energy over pretrained directions.}
To capture where the update injects energy, we project the effective update $\Delta W=W^{*}-W_{0}$ onto pretrained input directions:
\begin{equation}
  E_{\Delta}(i) = \lVert \Delta W v_{i}\rVert_{2}.
\end{equation}
Compared with the diagonal projection, $E_{\Delta}(i)$ captures both scaling changes and off-diagonal rotations along the $i$-th latent direction. We use this profile as a description of how different PEFT parameterizations allocate update energy, not as a standalone explanation of target-domain gains.

\textbf{Descriptive spectral smoothness}.
\autoref{fig:spectral_analysis} visualizes the retention and update-energy profiles. We summarize local irregularity with a fluctuation score, defined in Appendix~\ref{app:full-spectral-profiles}, and use it only as a descriptive measure rather than a formal taxonomy. The main pattern is that PiSSA and MiSS show large retention-side deviations, LoRA exhibits spiky update-energy allocation, and OFT maintains a more structured retention profile under its orthogonal parameterization. Full spectral profiles and smoothness statistics are provided in Appendix~\ref{app:full-spectral-profiles}; additional OFT-specific singular-vector diagnostics are provided in Appendix~\ref{app:oft-sva}.

\textbf{Capability-conditioned drift}.
The spectral profiles describe where an update acts, but not whether those directions are used by a data distribution. We therefore compute the following quantity:
\begin{equation}
\mathrm{CSD}_{D}
=
\mathbb{E}_{x\sim D}
\frac{\|\Delta W h_0(x)\|_2^2}
{\|W_0 h_0(x)\|_2^2+\epsilon}.
\end{equation}
where $h_0(x)$ is the pretrained activation. Intuitively, $\mathrm{CSD}_D$ weights update energy by how strongly dataset $D$ activates the corresponding directions. We find that $\mathrm{CSD}_G$ is associated with forgetting, while $\mathrm{CSD}_T$ is not a simple predictor of target gain. Since CSD measures raw displacement and does not distinguish rotation-like movement from non-isometric distortion, we use it as a bridge from weight profiles to the activation-geometry analysis below, and full results are in Appendix~\ref{app:csd}.

\subsection{Activation-Space Geometry}
\label{sec:activation-geometry}

Weight-space geometry does not directly tell us whether an update damages concrete capabilities. We therefore compare base-model activations and finetuned-model activations on the same general examples for Qwen2.5-7B and Llama3.2-3B-Instruct checkpoints from the main table. We collect full-forward module outputs on general data (IFEval, NQ, BBH); the tested layer/module locations and full breakdown are reported in Appendix~\ref{app:activation-geometry}.

We use three complementary diagnostics. First, Procrustes residual removes the best shared orthogonal alignment between centered base activations $X_0$ and finetuned activations $X_1$:
\begin{equation}
d_{\mathrm{proc}} =
\frac{\min_{R^\top R=I}\|X_1R-X_0\|_F}{\|X_0\|_F+\epsilon}.
\end{equation}
A large residual indicates non-isometric distortion beyond a benign rotation. Second, linear CKA~\cite{kornblith2019similarity} measures representation similarity through centered Gram matrices. Third, pairwise Gram distortion compares the cosine-similarity structure among examples and is insensitive to a shared orthogonal rotation. Full definitions and detailed metrics are given in Appendix~\ref{app:activation-metrics}.

\begin{table}[t]
  \centering
  \footnotesize
  \setlength{\abovecaptionskip}{3.5pt}
  \setlength{\belowcaptionskip}{2pt}
  \setlength{\tabcolsep}{2.25pt}
  \renewcommand{\arraystretch}{1.18}
  \begin{tabular}{lccc}
    \textbf{Metric} & \textbf{External metric} & \textbf{Pearson} & \textbf{Spearman} \\
    \shline
    Procrustes residual & Forgetting & 0.711 & 0.568 \\
    Gram distortion & Forgetting & 0.485 & 0.361 \\
    CKA & Forgetting & -0.761 & -0.711 \\
  \end{tabular}
  \caption{Activation-geometry correlations between general-data representation geometry and forgetting across 20 SFT checkpoints of representative methods and eight module locations. Procrustes residual and Gram distortion measure non-isometric distortion, while CKA measures representation similarity.}
  \label{tab:activation-corr}
  \vspace{-3.5mm}
\end{table}

\begin{table}[t]
  \centering
  \footnotesize
  \setlength{\abovecaptionskip}{4pt}
  \setlength{\belowcaptionskip}{2pt}
  \setlength{\tabcolsep}{6.4pt}
  \renewcommand{\arraystretch}{1.16}
  \begin{tabular}{lcccc}
    \textbf{Method} & \textbf{Proc.} $\downarrow$ & \textbf{Gram} $\downarrow$ & \textbf{CKA} $\uparrow$ & \textbf{Forget} $\downarrow$ \\
    \shline
    Full FT & 0.1640 & 0.2500 & 0.8654 & 17.31 \\
    LoRA    & 0.1808 & 0.2430 & 0.8564 & 15.97 \\
    OFT     & \textbf{0.1279} & \textbf{0.1906} & \textbf{0.9340} & \textbf{7.81} \\
    MiLoRA  & 0.1635 & 0.2476 & 0.8651 & 16.35 \\
    PiSSA   & 0.4376 & 0.8655 & 0.4402 & 34.56 \\
  \end{tabular}
  \caption{Activation-geometry on general data for one representative module slice over SFT checkpoints. We averge the results for each method, and the complete layer/module breakdown is in Appendix~\ref{app:activation-layer-module}. Procrustes residual and Gram distortion measure non-isometric distortion, while CKA measures representation similarity.}
  \label{tab:activation-geometry}
  \vspace{-4mm}
\end{table}

\autoref{tab:activation-corr} summarizes the main correlations over general-distribution activation rows. Procrustes residual on general data strongly correlates with forgetting. Linear CKA shows the complementary trend, while pairwise Gram distortion also supports the relational-geometry interpretation with a weaker correlation.

\autoref{tab:activation-geometry} compares the activation geometry patterns across different PEFT methods. OFT exhibits lower non-isometric distortion and higher CKA than LoRA and full fine-tuning, whereas PiSSA emerges as a clear outlier, showing the strongest distortion and the most severe forgetting. This suggests that OFT’s advantage in general-capability retention is reflected not only in the geometry of the weights, but also in the functional geometry of the representations that support general capabilities.

These metrics are used as retention-side diagnostics. We do not use them to explain target-task gains, since plasticity on reasoning-heavy math and medical tasks may depend on task-aligned computation, answer margins, and multi-step reasoning behavior beyond representation geometry.

Taken together, these two internal perspectives clarify why retention varies across PEFT methods. Weight-space profiles characterize the update bias induced by each parameterization, while activation-space diagnostics indicate whether the resulting representations used for general evaluation remain geometrically stable. OFT better preserves this geometry, whereas PiSSA substantially distorts it.

\vspace{-1.25mm}
\begin{mdframed}
[leftmargin=0.2em, rightmargin=0.2em]
\fontsize{10pt}{\baselineskip}
\itshape
\textbf{Takeaway}.
PEFT's general-capability retention is strongly associated with preserving the representation geometry of general-evaluation data.
\end{mdframed}
\vspace{-3mm}

\section{Interpolation as a Pathwise Study}
\label{sec:interpolation}
\vspace{-0.5mm}
The analyses above compare the base and final fine-tuned models, but the final checkpoint represents only a single point along the adaptation trajectory. A method may acquire most of its target-domain gains before reaching this endpoint, while continued movement along the trajectory can further erode general capabilities.
This motivates a pathwise question: has the final checkpoint moved farther than necessary to achieve its target-task gains?
We use interpolation to probe this question. Rather than treating interpolation as a new evaluation protocol or adaptation technique, we use it as a diagnostic tool to trace how target performance and general retention change as the model moves from the base checkpoint toward the adapted checkpoint.

\begin{figure}[t!]
  \centering
  \setlength{\abovecaptionskip}{5pt}
  \setlength{\belowcaptionskip}{5pt}
  \includegraphics[width=\linewidth]{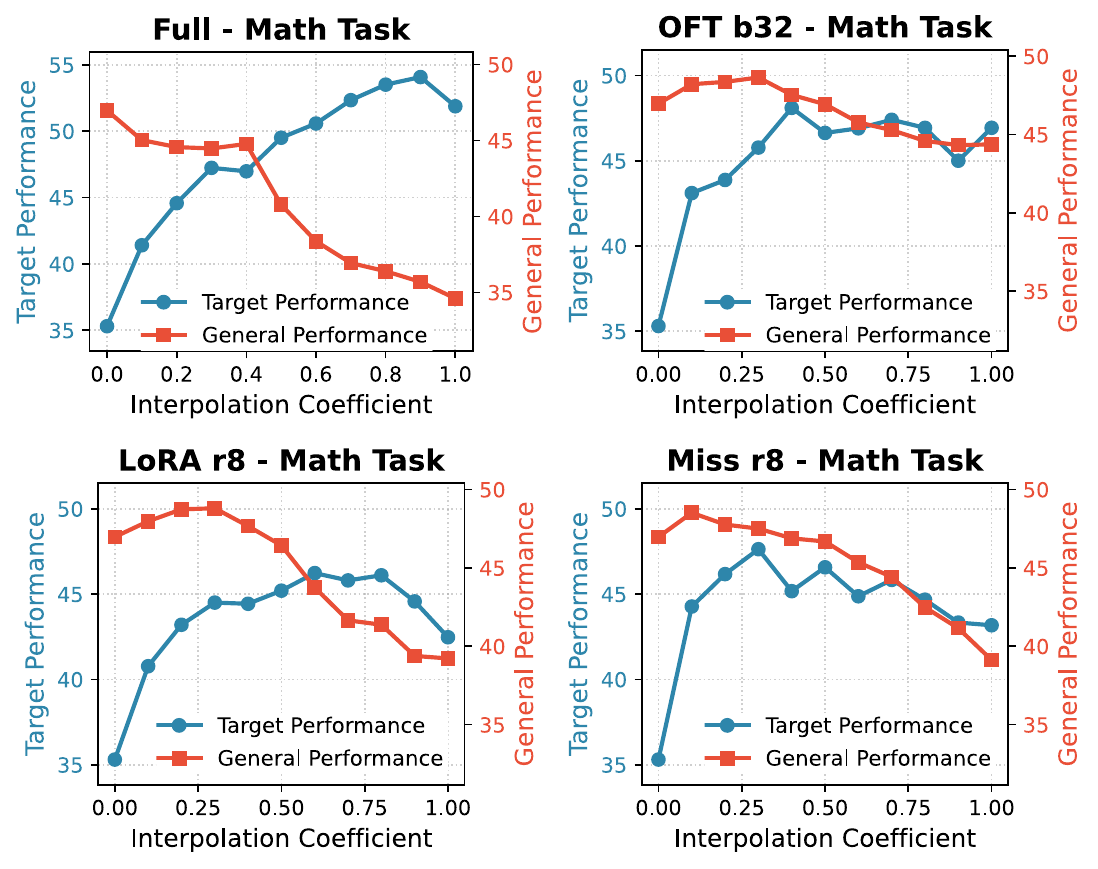}
  \caption{Target-general performance trade-off with $\alpha$-interpolation for PEFT methods. Interpolation is used as a pathwise diagnostic of SFT overshoot. Additive methods are interpolated by scaling $\Delta W$; OFT is interpreted through its parameterization-aware Cayley path.}
  \label{fig:interpolate-dualaxes}
  \vspace{-1em}
\end{figure}

\subsection{Finetuning Overshoots the Best Trade-off}
\label{sec:interpolation:sft-overshoot}

For full finetuning and additive PEFT methods, the natural path scales the effective update $\Delta W$. Sweeping this coefficient exposes a target-retention curve. In SFT, we consistently observe that the final checkpoint lies past a better operating point, and reducing the update strength can recover general ability while preserving much of the target gain.

Let $W_{0}$ be the pretrained weight and $W^{*}$ be the finetuned effective weight. For full finetuning and additive PEFT methods, the natural interpolation is
\begin{equation}\label{eq:linear_intep}
  W(\alpha)= W_{0}+ \alpha\,(W^{*}-W_{0}),\quad \alpha\in[0,1],
\end{equation}
where $\alpha=0$ recovers the base model and $\alpha=1$ recovers the final adapted model. Sweeping $\alpha$ exposes a pathwise stability-plasticity curve.

\autoref{fig:interpolate-dualaxes} shows that SFT checkpoints often exhibit an overshoot phenomenon: the final checkpoint is not always the best target-retention operating point. Moving back along the interpolation path can recover general ability while preserving much of the target gain. This is not meant to replace final-checkpoint evaluation; instead, it diagnoses where stability is lost along the adaptation path.

\begin{figure}[t]
  \centering
  \setlength{\abovecaptionskip}{4pt}
  \setlength{\belowcaptionskip}{2pt}
  \includegraphics[width=.95\linewidth]{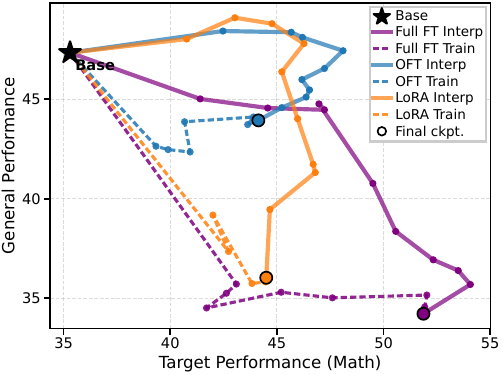}
  \caption{Training trajectory versus interpolation trajectory in the target-general plane. Dotted curves follow saved optimization checkpoints, while solid curves sweep $\alpha$ along the interpolation path between the base model and the final checkpoint. }
  \label{fig:train-vs-interp}
  \vspace{-3mm}
\end{figure}

\autoref{fig:train-vs-interp} further compares the actual training trajectory with the interpolation trajectory. The two paths exhibit totally different patterns, with the training trajectory being concave and the interpolation trajectory being convex. This distinction suggests that the SFT overshoot phenomenon is not merely a consequence of late-stage overfitting, and that the optimal stability-plasticity trade-off cannot be attained through early stopping. We have also discussed a similar overshoot pattern of longer-horizon RLVR high-$k$ degradation in Appendix~\ref{app:rlvr-highk-interp}.

\subsection{Parameterization-Aware Interpolation}
\label{sec:interpolation:parameterization-aware}

Different PEFT parameterizations have different natural interpolation coordinates. For additive methods, scaling $\Delta W$ gives the natural path. For OFT, adaptation is parameterized by a skew-symmetric generator $Q$ and an orthogonal Cayley transform, so dense-weight interpolation leaves the OFT manifold. We therefore interpolate by scaling the generator, $R(\alpha)=\mathrm{Cayley}(\sqrt{\alpha}Q)$, where $\alpha$ controls rotation strength. This path preserves the intended orthogonal geometry; dense-weight interpolation is used only as a diagnostic baseline.

We empirically find that this distinction matters. On Qwen2.5-7B SFT-math with OFT-b32, the $\sqrt{\alpha}Q$ path gives a better target-retention frontier than dense-weight delta interpolation: at $\alpha=0.3$, it reaches 45.77 math and 48.64 General, while the linear interpolation (as in Equation~\ref{eq:linear_intep}) at the same $\alpha$ gives 43.93 math and 43.91 General. Implementation details and the complete interpolation comparison are reported in Appendix~\ref{app:interpolation-setup},\ref{app:oft-cayley}.

\subsection{Layer-wise Interpolation for OFT}
\label{sec:interpolation:alt}

The parameterization-aware view also prompts us to reconsider whether a single global interpolation coefficient is too coarse. We observe that OFT updates are imbalanced across depth, with later layers often receiving substantially larger rotation strengths than early layers (see the evidence in \autoref{fig:oft-layer-update-strength} of Appendix~\ref{app:layerwise-rewinding}); uniform interpolation rewinds all layers using the same coefficient and therefore cannot correct this layer-wise imbalance.

To further investigate this, we perform a simple layer-wise rewinding experiment. Instead of applying one coefficient to all layers, we rescale each layer's generator $Q_{\ell}$ separately while staying on the OFT manifold. \textbf{SafeScale} uses the empirical average update strength of the first five layers (for Qwen2.5-7B) as the reference and scales other layers toward it, while \textbf{MinScale} uses the weakest-updated layer as the reference. As shown in \autoref{tab:ablation-trade-off}, layer-wise rewinding improves the target-retention trade-off relative to the final OFT checkpoint. These results suggest that SFT overshoot is not only related to the total update magnitude, but also to how adaptation strength is allocated across layers. 
The improvement from this sweeping-free adjustment indicates that adaptively choosing rewinding strengths for different layers is a promising direction for post-hoc trade-off control. 
Additional results for OFT and LoRA/MiSS variants are reported in Appendix~\ref{app:layerwise-rewinding},\ref{app:additive-rewinding}.
\begin{table}[t]
  \centering
  \footnotesize
  \setlength{\abovecaptionskip}{3pt}
  \setlength{\belowcaptionskip}{3pt}
  \setlength{\tabcolsep}{2pt}
  \renewcommand{\arraystretch}{1.285}
  \resizebox{\linewidth}{!}{%
  \begin{tabular}{lcccc}
    \textbf{Variant}           & \textbf{Target (Math)} & \textbf{General (Math)} & \textbf{Target (Med)} & \textbf{General (Med)} \\
    \shline Finetuned          & 46.93 (+11.63)        & 44.37 (-2.60)          & 48.63 (+2.27)        & 42.40 (-4.57)         \\
    Uniform scal. 0.8        & 46.93 (+11.63)        & 44.82 (-2.15)          & 49.17 (+2.81)        & 45.76 (-1.21)         \\
    Uniform scal. 0.4        & 48.10 (+12.80)        & 47.53 (+0.56)          & 48.83 (+2.47)        & 48.15 (+1.18)         \\
    Layer. scal. (Safe) & 47.17 (+11.87)        & 46.69 (-0.28)          & 50.01 (+3.65)        & 47.61 (+0.64)         \\
    Layer. scal. (Min)  & 47.83 (+12.53)        & 46.86 (-0.11)          & 49.76 (+3.41)        & 47.79 (+0.82)         \\
  \end{tabular}
  }
  \caption{OFT post-hoc rewinding case study. Layer-wise generator scaling can recover general ability while preserving much of the target gain.}
  \label{tab:ablation-trade-off}
  \vspace{-3.5mm}
\end{table}

\section{Concluding Remarks}
\vspace{-0.35mm}

We introduce PEFT-Arena to evaluate PEFT methods through the stability-plasticity trade-off rather than target accuracy alone. Parameterizations differ substantially in how they exchange target gains for general-capability retention, with OFT often defining a strong frontier at comparable parameter budgets.
These differences are reflected in model geometry. Weight-space profiles expose how each parameterization interacts with pretrained spectral structure, while activation-space diagnostics tie forgetting to non-isometric distortion of general-evaluation representations. Raw movement is therefore an insufficient proxy: OFT shifts representations while preserving their relational geometry, whereas PiSSA and full finetuning induce stronger distortion.
Interpolation, used as a pathwise diagnostic, further shows that final SFT checkpoints can drift past a better target-retention operating point, and that interpolation paths should respect each method's update geometry. PEFT methods should thus be designed not only for target ability acquired, but for the pretrained geometry preserved.

\newpage
\section*{Limitations}

This work studies PEFT methods through the stability-plasticity trade-off, and its scope suggests several directions for future work. PEFT-Arena focuses on two reasoning-oriented target domains, mathematics and medicine, and evaluates retention mainly with English general-evaluation suites. Broader multilingual, dialogue, and safety-oriented evaluations would further extend the benchmark.

Our method coverage emphasizes weight-parameterized PEFT methods such as LoRA variants, OFT, IA$^3$, and related baselines. Prompt-, prefix-, and adapter-based methods are also important PEFT families, but their adaptation mechanisms are less directly expressed as explicit weight perturbations. Extending the same evaluation and analogous activation-space diagnostics to these families would make the benchmark more comprehensive.

The internal analyses are empirical diagnostics rather than a complete causal theory of forgetting. They are strongest for characterizing retention-side representation changes, while target-domain plasticity, especially for reasoning-heavy tasks, may require additional task-aligned diagnostics. Similarly, interpolation is used as a pathwise diagnostic and post-hoc trade-off control, with SFT providing the clearest target--retention curves and longer RLVR providing complementary high-$k$ evidence.

\bibliography{reference}

\clearpage
\newpage
\onecolumn

\addcontentsline{toc}{section}{Appendix} %
\renewcommand \thepart{} %
\renewcommand \partname{}
\part{\Large{\centerline{Appendix}}}
\parttoc

\newpage

\appendix
\section{Related Work}
\label{app:related-work}

\textbf{PEFT methods}. Parameter-efficient finetuning (PEFT) specializes large pretrained models by optimizing only a small subset of parameters, substantially reducing computational overhead while often achieving performance comparable to full finetuning~\cite{houlsby2019parameter,aghajanyan2020intrinsic,hu2022lora,edalati2022krona,wang2022adamix,gheini2021cross,zaken2022bitfit,guo2020parameter,sung2021training,ansell2022composable,lester2021power,li2021prefix,vu2022spot,he2021towards,mao2021unipelt,karimi2021compacter,liu2022few,sung2022lst,chen2022parameter,jia2022visual,chen2022adaptformer,zhang2022neural,jie2023fact,lian2022scaling,luo2023towards}. Prior PEFT methods can be broadly grouped into three categories.
\textit{(1) Additive weight reparameterization.} A representative approach is Low-Rank Adaptation (LoRA)~\cite{hu2022lora}, which freezes pretrained weights and introduces a learnable low-rank update to enable efficient task adaptation with a small number of trainable parameters.
Subsequent work extends LoRA along several complementary dimensions. 
Methods such as AdaLoRA~\cite{zhang2023adaptive} focus on adaptive rank allocation by dynamically re-allocating rank budgets during training, and DoRA~\cite{liu2024dora} improves optimization by decoupling update magnitude from direction. 
Other approaches modify the update structure: MISS~\cite{kang2025missrevisitingtradeofflora} replaces the two-factor decomposition with a single expanded low-rank matrix, and VeRA~\cite{kopiczko2024vera} attains high effective update rank through shared, frozen random projection matrices modulated by lightweight trainable scaling vectors.
Another line of work explores initialization strategies informed by the spectral properties of pretrained weights. PiSSA~\cite{meng_pissa_2024} initializes low-rank factors using principal singular components, whereas MiLoRA~\cite{wang2025miloraharnessingminorsingular} instead targets minor components by initializing updates on the minor singular subspace while freezing the principal components.
\textit{(2) Constrained weight updates.}
Beyond low-rank additive updates, Orthogonal Finetuning (OFT)~\cite{qiu2023controlling,liu_parameter-efficient_2024,qiu_orthogonal_2025} constrains the learned transformations to be orthogonal. By preserving the spectral properties of the original weight matrices, such constraints act as an implicit regularizer during adaptation and provide an alternative form of structured parameter efficiency.
\textit{(3) Activation- and prompt-based adaptation.} Instead of directly modifying weight matrices, this line of work introduces auxiliary parameters to modulate intermediate representations. For instance, IA$^3$ learns per-channel scaling vectors that element-wise modulate key activations~\cite{liu2022few}. Similarly, Prompt Tuning~\cite{liu2022p} and Prefix-Tuning~\cite{li2021prefix} prepend trainable continuous embeddings to the input layer or to hidden states across attention blocks, respectively. These methods facilitate task specialization by re-contextualizing frozen representations, effectively inducing localized shifts in the activation space.

\textbf{Continual learning}. Continual learning (CL) aims to incorporate new knowledge while preserving previously acquired capabilities, thereby mitigating catastrophic forgetting \cite{wang2024comprehensive}. Prior CL approaches are commonly categorized into data-replay-based methods \cite{aljundi2019gradient,silver2002task,tiwari2022gcr}, which require access to historical training data, and data-free methods \cite{wortsman2022robust,lubana2022quadratic,chen2024mofo,panda2024lottery}, which avoid replay by relying instead on architectural constraints or regularization. These data-free CL methods are relevant to our setting because they aim to preserve prior capabilities without replaying the original pretraining data.
This paradigm operates directly in parameter space, bypassing the need for historical data. Standard approaches employ $\ell_2$ regularization \cite{kirkpatrick2017overcoming} or model merging \cite{wortsman2022robust,lubana2022quadratic,kleiman2025soup,lin2024mitigating} to anchor finetuned parameters to the base model, thereby balancing stability and plasticity. For PEFT, O-LoRA \cite{wang2023orthogonal} and InfLoRA \cite{liang2024inflora} enforce orthogonality between sequential task-specific subspaces to mitigate forgetting. However, these methods are designed mainly for sequential continual-learning protocols, whereas PEFT-Arena studies one-shot post-training and evaluates retention of pretrained capabilities after a single adaptation run. KeepLoRA \cite{luo2026keeplora} constrains updates to the non-principal subspace of the original weights, protecting the principal components that encode general capabilities.

\textbf{Catastrophic forgetting in LLMs}.
While traditional CL primarily addresses multi-task sequential learning, recent focus has shifted toward catastrophic forgetting during the post-training of LLMs, particularly when finetuning on distributions that diverge significantly from the pre-training data \cite{sanyal2025upweighting,lin2025sft}. A prevailing hypothesis identifies the discrepancy between on-policy and off-policy distributions as a primary driver of forgetting. To bridge this gap, several data-centric strategies aim to simulate on-policy learning via reweighting. FLOW \cite{sanyal2025upweighting} prioritizes simpler samples to stabilize updates, while TALR \cite{lin2025sft} and EAFT \cite{diao2026entropy} mitigate forgetting by attenuating learning rates or weights for tokens with high difficulty or low information entropy.
These methods are complementary to our study, focusing at the data, loss, or optimizer level, while PEFT-Arena focuses on comparing PEFT parameterizations and analyzing how their parameter-space and activation-space geometry lead to different stability-plasticity trade-offs.
Despite the effectiveness of these data-level interventions, a fundamental question remains: how do these stability-promoting effects manifest in the model’s underlying parameter space? Emerging evidence in reinforcement learning suggests that optimization stability is intrinsically linked to parameter-space dynamics. 
Recent studies also suggest that retention can be linked to parameter-space dynamics, such as preferentially updating non-principal weight components~\cite{zhu2025path,mukherjee2025reinforcement}, and that LoRA may retain pretrained knowledge better than full finetuning~\cite{biderman2024lora}. PEFT-Arena builds on these observations by evaluating multiple PEFT families under a unified protocol and by connecting external trade-offs to internal geometry diagnostics.

\textbf{Model averaging and interpolation}.
Task arithmetic~\cite{ilharco2023editing} shows that meaningful task-specific knowledge is encoded in weight differences $W_{\text{ft}} - W_{\text{pre}}$, and that scaling or combining these ``task vectors'' enables modular control over model behavior. Model soups~\cite{wortsman2022robust} average checkpoints for improved robustness. WiSE-FT~\cite{wortsman2022robust} interpolates between zero-shot and finetuned weights to improve robustness under distribution shift while maintaining target performance. TIES merging~\cite{yadav2023tiesmerging} resolves sign conflicts across task vectors. Fisher merging~\cite{matena2022merging} uses second-order information for weighted combinations. Orthogonal merging~\cite{yang2026orthogonal} combines task-specific weights on orthogonal manifolds. Functional dual anchors~\cite{shi2025model} merge task vectors within input-representation space. These methods primarily combine multiple task-specialized models. 
Our use of interpolation is related in form but different in purpose. 
Rather than proposing a new model-merging technique, merging multiple task-specialized models, or targeting distribution-shift robustness, we use interpolation as a pathwise diagnostic for single-task PEFT. 
By sweeping the interpolation coefficient $\alpha$ between the base and finetuned model, we trace explicit stability-plasticity trade-off curves and diagnose when final SFT checkpoints overshoot a better target-retention operating point. We further emphasize that the interpolation path should respect the natural coordinates of each PEFT parameterization.

\textbf{Spectral analysis of neural networks}.
The singular value spectrum of weight matrices has long been recognized as informative about network behavior. Heavy-tail spectral theory connects weight matrix spectra to generalization~\cite{martin2021implicit}, and recent work in reinforcement learning has linked stability to preferential updating of non-principal weight components~\cite{zhu2025path,mukherjee2025reinforcement}. \cite{biderman2024lora} empirically observes that LoRA retains pretrained knowledge better than full finetuning, but the spectral mechanisms underlying this observation remain unexplored. \cite{liu2021orthogonal,qiu2023controlling} explicitly link the pretrained knowledge retention to spectrum preservation. Our spectral analysis specifically targets PEFT updates, decomposing changes into retention and adaptation profiles and using them as descriptive diagnostics of update geometry. These weight-space diagnostics are complemented by activation-space analysis, which measures whether finetuning preserves representations on general-evaluation data. This offers a systematic spectral framework that connects internal geometric properties of PEFT updates to external stability-plasticity outcomes.

\textbf{Representation similarity and activation geometry}.
Neural representation similarity has been widely studied using CCA-based metrics, SVCCA, PWCCA, and CKA~\cite{kornblith2019similarity,raghu2017svcca,morcos2018insights}. CKA compares centered Gram matrices and is now a standard diagnostic for representation similarity~\cite{kornblith2019similarity}, while recent benchmarking studies show that orthogonal Procrustes is a strong and interpretable baseline for comparing neural representations~\cite{ding2021grounding}. 

This perspective is also relevant to forgetting: prior work studies catastrophic forgetting through hidden representation changes and shows that forgetting can manifest as representation drift, often in deeper layers~\cite{ramasesh2021anatomy,davari2022probing}. Continual-learning studies further argue that reducing representation drift on previously learned data can mitigate forgetting~\cite{caccia2021reducing}. Motivated by these works, we analyze PEFT-induced changes in general-capability activations. In particular, we distinguish raw movement from non-isometric distortion: orthogonal transformations preserve inner products, distances, and angles, so large activation movement is not necessarily destructive. We therefore use Procrustes residual, pairwise Gram distortion, and CKA to measure whether finetuning preserves the relational geometry of pretrained representations.

\section{Implementation and Evaluation Details}
\label{app:implementation}

\subsection{Model and Training Details}
\label{app:model-training-details}

\textbf{Compute.} All experiments are conducted on 8$\times$ NVIDIA H100 80GB GPUs, each by a single run. The total compute varies by model, method, and training regime; for instance, a typical LoRA SFT run (Math / Medical training, target and general evaluation) takes approximately 100 GPU-hours for 7B models and 60 GPU-hours for 3B models. A 200-step RLVR runs require approximately 200 GPU-hours. Diagnostic analyses are run offline and do not require additional model training.

\textbf{SFT.} We use an effective batch size of 256, a maximum response length of 8192 tokens, and train for 4 epochs.
Full fine-tuning uses a learning rate of $5\times10^{-5}$, while all PEFT methods use $2\times10^{-4}$.
We adopt a cosine decay learning-rate scheduler.

\textbf{RL.} We use a rollout batch size of 256, a mini-batch size of 64 for gradient accumulation, and a group size of 8.
The maximum generation length is set to 8192 tokens, consistent with SFT and evaluation.
Full fine-tuning uses a learning rate of $10^{-6}$, while all PEFT methods use $10^{-5}$.
Unless otherwise specified, RL experiments are trained for 200 steps. For the longer-training study in \autoref{sec:rl-longer-training}, we additionally train to 500 steps.

\subsection{Evaluation Protocol}
\label{app:evaluation-protocol}

We elaborate the evaluation settings for tasks of different domains below.

\textbf{Math.} Each math problem is evaluated by average accuracy@16, with a maximum response length of 8192 tokens (consistent with training) and temperature $T=0.6$.

\textbf{Medical.} We use average accuracy as the primary metric with temperature $T=0.0$.

\textbf{General.} We follow the evaluation configuration in OpenCompass\footnote{https://github.com/open-compass/opencompass} with context length 1024, temperature $T=0.0$, and one sample per query.

\subsection{Artifacts, Licenses, and Reproducibility}
\label{app:artifacts}

We use publicly available model checkpoints, PEFT implementations, evaluation datasets, and benchmark frameworks, and cite their creators in the main text and related work. We use these artifacts under their respective licenses and usage terms. The evaluated models, trainable parameter counts, PEFT configurations, training hyperparameters, decoding settings, context length, and evaluation protocol are described in this appendix. We provide code and configurations at \url{https://github.com/Sphere-AI-Lab/PEFT-Arena} to support reproducibility, subject to the licenses and terms of the underlying artifacts. We do not collect new user data or human-subject data. For public benchmark datasets, we rely on their original curation and use them only for controlled research evaluation.

\section{Expanded Benchmark Results}
\label{app:expanded-benchmarks}

\subsection{Expanded General-Capability Benchmarks}
\label{app:expanded-general}

To test whether the original General axis transfers beyond IFEval/NQ/BBH, we further evaluate representative checkpoints on HumanEval, HellaSwag, WinoGrande, MMLU(avg), ARC, and GSM8K. The expanded benchmarks are used as consistency checks for the General axis rather than as a separate main benchmark suite. The detailed per-task scores are reported in \autoref{tab:app:expanded-general-results}.

\begin{table}[htbp]
  \centering
  \scriptsize
  \setlength{\abovecaptionskip}{4pt}
  \setlength{\belowcaptionskip}{8pt}
  \setlength{\tabcolsep}{11pt}
  \renewcommand{\arraystretch}{1.25}
  \begin{adjustbox}{max width=\linewidth}
  \begin{tabular}{l|cccccc|c}
    Method & HumanE. & Hella. & Wino. & MMLU & ARC & GSM8K & Avg. \\
    \shline
    \multicolumn{8}{l}{\textbf{A: Llama, RL-Math}} \\
    Full FT      & 58.54 & 64.40 & 56.99 & 62.38 & 76.95 & 79.15 & 66.40 \\
    OFT-b32      & 59.76 & 64.75 & 53.91 & 62.07 & 75.59 & 78.47 & 65.76 \\
    LoRA-r8      & 59.76 & 64.17 & 55.72 & 62.25 & 77.63 & 77.71 & 66.21 \\
    \multicolumn{8}{l}{\textbf{B: Llama, SFT-Math}} \\
    Full FT      & 54.88 & 63.77 & 12.71 & 59.83 & 41.69 & 71.80 & 50.78 \\
    OFT-b16      & 55.49 & 60.97 & 22.89 & 59.78 & 65.76 & 75.82 & 56.79 \\
    OFT-b128     & 54.88 & 60.05 & 15.47 & 57.99 & 50.17 & 70.28 & 51.47 \\
    LoRA-r4      & 54.88 & 63.18 & 11.60 & 56.82 & 41.02 & 72.48 & 50.00 \\
    LoRA-r8      & 48.78 & 63.54 & 12.23 & 58.41 & 39.66 & 73.54 & 49.36 \\
    LoRA-r16     & 35.98 & 60.09 & 12.31 & 58.37 & 40.34 & 71.34 & 46.41 \\
    LoRA-r32     & 40.85 & 60.71 & 14.84 & 58.87 & 39.32 & 69.98 & 47.43 \\
    AdaLoRA-r8   & 54.88 & 62.30 & 13.34 & 58.66 & 41.02 & 62.32 & 48.75 \\
    PiSSA-r8     & 0.00  & 24.79 & 4.89  & 25.18 & 12.54 & 1.97  & 11.56 \\
    MiSS-r8      & 49.39 & 57.06 & 24.63 & 57.76 & 47.46 & 69.75 & 51.01 \\
    MiSS-r64     & 39.02 & 53.67 & 23.91 & 55.65 & 41.36 & 69.90 & 47.25 \\
    VeRA-r256    & 56.10 & 62.08 & 57.54 & 62.24 & 71.86 & 77.48 & 64.55 \\
    DoRA-r8      & 39.63 & 63.03 & 13.26 & 57.97 & 39.32 & 71.65 & 47.48 \\
    IA$^3$       & 54.27 & 57.74 & 54.85 & 61.45 & 69.49 & 76.19 & 62.33 \\
    \multicolumn{8}{l}{\textbf{C: Llama, SFT-Med}} \\
    Full FT      & 48.78 & 57.33 & 17.13 & 58.76 & 40.34 & 72.71 & 49.18 \\
    OFT-b16      & 51.83 & 61.49 & 13.89 & 58.36 & 60.68 & 76.35 & 53.77 \\
    OFT-b32      & 51.83 & 60.04 & 15.15 & 58.13 & 55.59 & 74.75 & 52.58 \\
    OFT-b64      & 45.12 & 58.98 & 14.21 & 59.13 & 52.20 & 74.75 & 50.73 \\
    OFT-b128     & 41.46 & 56.96 & 14.52 & 59.15 & 41.69 & 74.22 & 48.00 \\
    LoRA-r4      & 52.44 & 58.68 & 14.52 & 58.22 & 45.76 & 74.15 & 50.63 \\
    LoRA-r8      & 27.44 & 59.24 & 15.63 & 59.13 & 49.15 & 73.62 & 47.37 \\
    LoRA-r16     & 22.56 & 57.90 & 17.68 & 58.31 & 39.66 & 73.77 & 44.98 \\
    AdaLoRA-r8   & 51.22 & 59.04 & 13.26 & 56.90 & 46.10 & 74.98 & 50.25 \\
    MiSS-r8      & 47.56 & 60.22 & 16.10 & 56.20 & 44.41 & 71.27 & 49.29 \\
    MiSS-r64     & 41.46 & 53.42 & 18.47 & 53.87 & 32.54 & 66.34 & 44.35 \\
    DoRA-r8      & 33.54 & 59.60 & 14.84 & 58.52 & 49.49 & 74.60 & 48.43 \\
    IA$^3$       & 53.05 & 60.52 & 56.83 & 61.28 & 65.42 & 75.13 & 62.04 \\
    \multicolumn{8}{l}{\textbf{D: Qwen, RL-Math}} \\
    Full FT      & 76.83 & 84.88 & 61.01 & 52.89 & 29.32 & 83.44 & 64.73 \\
    OFT-b32      & 74.39 & 84.97 & 62.35 & 59.13 & 35.09 & 81.50 & 66.24 \\
    LoRA-r8      & 76.83 & 85.15 & 62.04 & 61.92 & 35.93 & 83.02 & 67.48 \\
    \multicolumn{8}{l}{\textbf{E: Qwen, SFT-Math}} \\
    Full FT      & 71.34 & 76.59 & 60.22 & 29.68 & 33.90 & 75.13 & 57.81 \\
    OFT-b16      & 70.73 & 82.14 & 60.54 & 30.93 & 37.63 & 83.32 & 60.88 \\
    OFT-b32      & 75.00 & 78.98 & 59.35 & 28.88 & 35.25 & 82.41 & 59.98 \\
    OFT-b64      & 75.61 & 82.27 & 61.80 & 29.11 & 34.58 & 81.65 & 60.84 \\
    OFT-b128     & 75.00 & 73.64 & 61.40 & 29.41 & 29.15 & 80.67 & 58.21 \\
    LoRA-r4      & 75.61 & 69.79 & 61.64 & 29.24 & 33.56 & 65.73 & 55.93 \\
    AdaLoRA-r8   & 75.00 & 64.20 & 60.54 & 29.39 & 34.92 & 57.24 & 53.55 \\
    PiSSA-r8     & 0.61  & 24.89 & 46.65 & 23.28 & 45.42 & 16.30 & 26.19 \\
    VeRA-r256    & 74.39 & 83.67 & 63.61 & 50.29 & 32.88 & 84.84 & 64.95 \\
    DoRA-r8      & 76.22 & 76.63 & 61.80 & 29.52 & 33.56 & 75.44 & 58.86 \\
    \multicolumn{8}{l}{\textbf{F: Qwen, SFT-Med}} \\
    Full FT      & 77.44 & 72.91 & 66.54 & 28.96 & 32.54 & 75.44 & 58.97 \\
    OFT-b32      & 77.44 & 70.61 & 65.59 & 28.67 & 32.88 & 85.06 & 60.04 \\
    OFT-b64      & 78.05 & 74.08 & 65.04 & 27.86 & 26.78 & 85.82 & 59.61 \\
    LoRA-r4      & 78.05 & 55.91 & 64.48 & 27.47 & 31.19 & 84.99 & 57.02 \\
    LoRA-r16     & 65.24 & 58.72 & 63.69 & 27.55 & 34.24 & 85.75 & 55.87 \\
    LoRA-r32     & 65.24 & 62.70 & 60.85 & 27.32 & 33.90 & 85.97 & 56.00 \\
    MiLoRA-r8    & 78.05 & 59.62 & 64.01 & 27.41 & 31.86 & 82.79 & 57.29 \\
    MiSS-r8      & 65.24 & 61.31 & 61.40 & 28.50 & 25.42 & 79.83 & 53.62 \\
    MiSS-r64     & 72.56 & 60.94 & 56.27 & 32.03 & 39.32 & 77.71 & 56.47 \\
    PiSSA-r8    & 17.68 & 24.91 & 45.94 & 9.54  & 50.17 & 39.58 & 31.30 \\
    VeRA-r256    & 71.95 & 83.37 & 61.17 & 57.85 & 33.90 & 80.59 & 64.81 \\
  \end{tabular}
  \end{adjustbox}
  \caption{Expanded general-capability raw results on additional benchmarks. We report HumanEval, HellaSwag, WinoGrande, MMLU(avg), ARC, and GSM8K for the same settings used in the main benchmark. For duplicated raw rows in Qwen2.5-7B RL-math (Full FT / OFT-b32), values are consolidated by averaging non-null duplicate entries after dropping missing values. ``--'' denotes unavailable evaluations.}
  \label{tab:app:expanded-general-results}
\end{table}

\subsection{Expanded Task-Level Scores for the Main Benchmark}
\label{app:detailed-main-results}

For completeness, we expand each average score in \autoref{tab:main-results} into its constituent task-level accuracies. Math target averages are computed over Math-500, AIME24, and AMC23 (\autoref{tab:app:detailed-math-scores}). Medical target averages are computed over the eleven medical benchmarks used in the main benchmark (\autoref{tab:app:detailed-med-scores}). General averages are computed over IFEval (Inst-level-strict-accuracy), NQ, and the BBH average (\autoref{tab:app:detailed-general-scores}).

\vspace{4mm}
\begin{table*}[h]
  \centering
  \footnotesize
  \setlength{\abovecaptionskip}{7pt}
  \setlength{\belowcaptionskip}{5pt}
  \setlength{\tabcolsep}{8pt}
  \renewcommand{\arraystretch}{1.25}
  \begin{tabular}{l|cccc|cccc}
     & \multicolumn{4}{c|}{Qwen2.5-7B-base} & \multicolumn{4}{c}{Llama3.2-3B-Instruct} \\
    Config & Math-500 & AIME24 & AMC23 & Avg. & Math-500 & AIME24 & AMC23 & Avg. \\
    \shline
    SFT/Base           & 61.30 & 7.10  & 37.50 & 35.30 & 49.20 & 9.80  & 24.40 & 27.80 \\
    SFT/Full FT        & 79.10 & 17.50 & 55.30 & 50.63 & 56.40 & 10.80 & 34.50 & 33.90 \\
    SFT/OFT-b16        & 71.30 & 11.20 & 44.50 & 42.33 & 52.30 & 6.00  & 30.00 & 29.43 \\
    SFT/OFT-b32        & 75.00 & 14.20 & 51.60 & 46.93 & 53.30 & 6.20  & 32.30 & 30.60 \\
    SFT/OFT-b64        & 74.70 & 14.60 & 49.40 & 46.23 & 53.30 & 6.00  & 28.60 & 29.30 \\
    SFT/OFT-b128       & 76.70 & 15.80 & 50.80 & 47.77 & 55.40 & 5.40  & 35.90 & 32.23 \\
    SFT/LoRA-r4a8      & 70.60 & 9.80  & 46.60 & 42.33 & 43.50 & 3.80  & 25.60 & 24.30 \\
    SFT/LoRA-r8a16     & 71.20 & 10.00 & 46.20 & 42.47 & 45.40 & 3.50  & 23.30 & 24.07 \\
    SFT/LoRA-r16a32    & 72.70 & 13.50 & 48.40 & 44.87 & 46.90 & 2.50  & 25.50 & 24.97 \\
    SFT/LoRA-r32a64    & 74.10 & 12.50 & 49.50 & 45.37 & 49.50 & 1.50  & 26.70 & 25.90 \\
    SFT/AdaLoRA-r8a16  & 68.50 & 10.00 & 42.80 & 40.43 & 41.40 & 0.60  & 20.50 & 20.83 \\
    SFT/PiSSA-r8a16    & 72.30 & 12.10 & 49.20 & 44.53 & 1.50  & 0.00  & 0.50  & 0.67  \\
    SFT/MiLoRA-r8a16   & 70.50 & 11.50 & 45.80 & 42.60 & 43.60 & 2.50  & 24.70 & 23.60 \\
    SFT/KeepLoRA-r8    & 68.00 & 9.80  & 43.80 & 40.53 & 30.30 & 2.30  & 13.00 & 15.20 \\
    SFT/MiSS-r8        & 71.00 & 12.10 & 46.40 & 43.17 & 44.00 & 0.80  & 25.30 & 23.37 \\
    SFT/MiSS-r64       & 75.90 & 13.80 & 51.10 & 46.93 & 52.10 & 2.70  & 31.10 & 28.63 \\
    SFT/VeRA-r256      & 67.70 & 7.50  & 43.10 & 39.43 & 51.60 & 9.20  & 25.60 & 28.80 \\
    SFT/DoRA-r8a16     & 71.80 & 10.20 & 45.00 & 42.33 & 44.50 & 1.50  & 25.50 & 23.83 \\
    SFT/IA3            & 61.80 & 5.80  & 37.80 & 35.13 & 52.20 & 6.90  & 30.00 & 29.70 \\
    RLVR/Full FT       & 76.10 & 13.90 & 52.70 & 47.57 & 52.80 & 10.40 & 26.20 & 29.80 \\
    RLVR/OFT-b32       & 74.40 & 16.50 & 52.80 & 47.90 & 51.00 & 10.00 & 28.90 & 29.97 \\
    RLVR/LoRA-r8a16    & 74.80 & 11.40 & 54.60 & 46.93 & 51.80 & 8.30  & 26.40 & 28.83 \\
  \end{tabular}
  \caption{Detailed math target scores for the checkpoints in \autoref{tab:main-results}.}
  \label{tab:app:detailed-math-scores}
\end{table*}

\begin{table*}[h!]
  \centering
  \small
  \setlength{\abovecaptionskip}{7pt}
  \setlength{\belowcaptionskip}{5pt}
  \setlength{\tabcolsep}{2pt}
  \renewcommand{\arraystretch}{1.25}
  \begin{adjustbox}{max width=\linewidth}
  \begin{tabular}{l|cccccccccccc}
    Config & GPQA & HLE & Lancet & MMLU-Pro & MB-op4 & MB-op5 & MedMCQA & MedQA & MedXpert & NEJM & PubMedQA & Avg. \\
    \shline
    SFT/Qwen/Base & 48.72 & 17.09 & 57.28 & 53.42 & 47.40 & 39.94 & 51.49 & 58.13 & 12.15 & 54.89 & 69.40 & 46.36 \\
    SFT/Llama/Base & 37.69 & 15.82 & 50.97 & 44.17 & 38.96 & 34.09 & 46.69 & 51.92 & 14.08 & 47.76 & 73.70 & 41.44 \\
    SFT/Qwen/Full FT & 47.44 & 12.66 & 61.65 & 64.43 & 59.09 & 57.79 & 60.79 & 70.46 & 16.77 & 62.52 & 76.30 & 53.63 \\
    SFT/Llama/Full FT & 36.92 & 10.76 & 53.64 & 45.41 & 49.03 & 45.13 & 51.45 & 59.54 & 14.01 & 52.24 & 68.70 & 44.26 \\
    SFT/Qwen/OFT-b16 & 37.69 & 13.29 & 54.37 & 54.98 & 52.27 & 39.94 & 54.20 & 60.17 & 12.97 & 55.72 & 72.30 & 46.17 \\
    SFT/Llama/OFT-b16 & 31.03 & 13.29 & 46.60 & 39.54 & 44.16 & 35.39 & 46.98 & 52.08 & 13.11 & 46.10 & 63.10 & 39.22 \\
    SFT/Qwen/OFT-b32 & 42.56 & 13.92 & 57.77 & 57.98 & 51.62 & 46.10 & 55.20 & 63.47 & 15.04 & 57.71 & 73.50 & 48.63 \\
    SFT/Llama/OFT-b32 & 31.79 & 10.13 & 47.09 & 39.41 & 44.16 & 37.01 & 46.40 & 52.47 & 12.15 & 46.43 & 67.50 & 39.50 \\
    SFT/Qwen/OFT-b64 & 40.51 & 8.86 & 59.71 & 61.11 & 55.52 & 46.43 & 56.06 & 67.01 & 16.22 & 59.20 & 73.50 & 49.47 \\
    SFT/Llama/OFT-b64 & 31.03 & 9.49 & 45.63 & 41.30 & 46.75 & 38.64 & 47.50 & 54.99 & 12.56 & 50.25 & 67.20 & 40.77 \\
    SFT/Qwen/OFT-b128 & 41.54 & 17.09 & 61.89 & 63.39 & 59.42 & 53.57 & 58.95 & 67.79 & 15.80 & 61.03 & 73.60 & 52.40 \\
    SFT/Llama/OFT-b128 & 35.38 & 8.86 & 48.79 & 43.13 & 45.45 & 43.18 & 49.18 & 56.17 & 12.42 & 49.92 & 68.20 & 42.17 \\
    SFT/Qwen/LoRA-r4a8 & 37.18 & 13.92 & 59.95 & 57.00 & 48.70 & 44.16 & 53.62 & 62.84 & 14.15 & 55.56 & 71.20 & 47.12 \\
    SFT/Llama/LoRA-r4a8 & 27.18 & 12.03 & 42.23 & 38.44 & 37.99 & 37.66 & 44.01 & 50.43 & 11.94 & 41.63 & 62.60 & 36.92 \\
    SFT/Qwen/LoRA-r8a16 & 42.56 & 14.56 & 57.77 & 57.26 & 50.32 & 46.75 & 54.58 & 62.22 & 13.87 & 55.39 & 71.70 & 47.91 \\
    SFT/Llama/LoRA-r8a16 & 29.74 & 12.03 & 46.12 & 40.52 & 39.61 & 34.74 & 45.49 & 51.61 & 12.77 & 42.62 & 66.50 & 38.34 \\
    SFT/Qwen/LoRA-r16a32 & 38.21 & 10.76 & 55.10 & 61.11 & 52.60 & 44.81 & 54.08 & 63.86 & 14.35 & 59.70 & 71.90 & 47.86 \\
    SFT/Llama/LoRA-r16a32 & 33.85 & 15.19 & 44.90 & 38.76 & 42.21 & 39.61 & 44.23 & 49.18 & 12.97 & 44.28 & 66.10 & 39.21 \\
    SFT/Qwen/LoRA-r32a64 & 40.77 & 11.39 & 55.34 & 60.52 & 58.77 & 47.08 & 55.25 & 64.89 & 16.70 & 59.54 & 74.00 & 49.48 \\
    SFT/Llama/LoRA-r32a64 & 25.38 & 12.66 & 50.24 & 38.96 & 45.78 & 37.99 & 46.12 & 50.04 & 11.94 & 48.09 & 65.40 & 39.33 \\
    SFT/Qwen/AdaLoRA-r8a16 & 37.95 & 13.92 & 54.37 & 54.40 & 44.81 & 41.88 & 52.47 & 60.96 & 14.29 & 54.23 & 68.20 & 45.22 \\
    SFT/Llama/AdaLoRA-r8a16 & 34.36 & 10.13 & 41.75 & 37.98 & 43.18 & 34.09 & 44.85 & 48.86 & 10.21 & 41.96 & 60.80 & 37.11 \\
    SFT/Qwen/PiSSA-r8a16 & 29.23 & 17.09 & 28.88 & 22.21 & 27.27 & 21.43 & 34.21 & 30.56 & 10.01 & 24.05 & 46.10 & 26.16 \\
    SFT/Llama/PiSSA-r8a16 & 20.00 & 16.46 & 25.24 & 16.03 & 25.32 & 19.81 & 26.03 & 25.53 & 10.56 & 21.89 & 26.00 & 21.17 \\
    SFT/Qwen/MiLoRA-r8a16 & 42.82 & 9.49 & 56.31 & 58.63 & 49.03 & 40.26 & 52.76 & 62.53 & 14.08 & 59.04 & 70.20 & 46.83 \\
    SFT/Llama/MiLoRA-r8a16 & 27.18 & 13.92 & 45.39 & 37.52 & 38.64 & 36.04 & 44.47 & 50.04 & 11.73 & 41.46 & 67.60 & 37.64 \\
    SFT/Qwen/KeepLoRA-r8 & 42.05 & 12.03 & 57.04 & 53.55 & 44.48 & 39.61 & 51.40 & 60.33 & 13.60 & 55.22 & 72.30 & 45.60 \\
    SFT/Llama/KeepLoRA-r8 & 37.18 & 18.99 & 48.06 & 39.67 & 44.16 & 40.58 & 47.33 & 54.52 & 11.80 & 43.62 & 67.90 & 41.26 \\
    SFT/Qwen/MiSS-r8 & 35.90 & 14.56 & 61.89 & 59.61 & 49.35 & 47.40 & 56.61 & 65.51 & 16.29 & 58.04 & 71.10 & 48.75 \\
    SFT/Llama/MiSS-r8 & 32.31 & 16.46 & 47.09 & 40.26 & 43.51 & 39.61 & 45.71 & 50.98 & 10.84 & 47.93 & 67.10 & 40.16 \\
    SFT/Qwen/MiSS-r64 & 45.90 & 12.66 & 58.74 & 62.08 & 57.79 & 53.25 & 58.69 & 67.95 & 17.12 & 61.19 & 75.50 & 51.90 \\
    SFT/Llama/MiSS-r64 & 39.49 & 9.49 & 46.60 & 39.80 & 44.81 & 45.78 & 50.18 & 54.99 & 13.39 & 48.59 & 68.40 & 41.96 \\
    SFT/Qwen/VeRA-r256 & 34.10 & 15.19 & 38.83 & 34.40 & 24.03 & 19.81 & 35.38 & 30.95 & 11.66 & 38.47 & 30.70 & 28.50 \\
    SFT/Llama/VeRA-r256 & 34.36 & 14.56 & 48.79 & 45.08 & 41.56 & 35.71 & 47.38 & 50.51 & 12.15 & 46.93 & 70.50 & 40.68 \\
    SFT/Qwen/DoRA-r8a16 & 38.46 & 14.56 & 58.01 & 56.87 & 56.49 & 43.18 & 53.45 & 64.96 & 13.25 & 58.54 & 70.70 & 48.04 \\
    SFT/Llama/DoRA-r8a16 & 26.67 & 14.56 & 43.20 & 38.63 & 40.91 & 38.96 & 45.16 & 49.02 & 11.87 & 44.61 & 67.20 & 38.25 \\
    SFT/Qwen/IA3 & 26.41 & 11.39 & 37.62 & 24.36 & 30.52 & 27.92 & 36.60 & 35.82 & 11.25 & 31.67 & 57.30 & 30.08 \\
    SFT/Llama/IA3 & 42.05 & 18.99 & 48.79 & 40.91 & 42.53 & 34.74 & 46.59 & 52.40 & 12.63 & 45.94 & 44.90 & 39.13 \\
    RLVR/Qwen/Full FT & 39.23 & 13.92 & 53.40 & 58.63 & 48.38 & 45.45 & 52.00 & 58.05 & 15.32 & 50.08 & 73.50 & 46.24 \\
    RLVR/Llama/Full FT & 42.31 & 14.56 & 50.73 & 50.23 & 48.70 & 45.45 & 53.93 & 58.60 & 13.18 & 51.74 & 75.20 & 45.88 \\
    RLVR/Qwen/OFT-b32 & 41.28 & 17.72 & 56.55 & 56.81 & 47.40 & 43.51 & 52.43 & 61.04 & 14.15 & 52.24 & 71.60 & 46.79 \\
    RLVR/Llama/OFT-b32 & 38.46 & 15.19 & 53.16 & 49.90 & 45.78 & 40.91 & 51.11 & 59.47 & 13.32 & 53.23 & 74.40 & 44.99 \\
    RLVR/Qwen/LoRA-r8a16 & 50.00 & 14.56 & 53.88 & 59.09 & 46.75 & 34.42 & 54.55 & 61.27 & 15.11 & 56.05 & 72.20 & 47.08 \\
    RLVR/Llama/LoRA-r8a16 & 47.69 & 12.66 & 50.97 & 46.78 & 48.38 & 45.13 & 50.44 & 60.33 & 13.53 & 49.92 & 68.90 & 44.97 \\
  \end{tabular}
  \end{adjustbox}
  \caption{Detailed medical target scores for the checkpoints in \autoref{tab:main-results}.}
  \label{tab:app:detailed-med-scores}
\end{table*}

\begin{table*}[h!]
  \centering
  \footnotesize
  \setlength{\abovecaptionskip}{7pt}
  \setlength{\belowcaptionskip}{-10pt}
  \setlength{\tabcolsep}{5pt}
  \renewcommand{\arraystretch}{1.25}
  \begin{adjustbox}{max width=\linewidth}
  \begin{tabular}{ll|cccc|cccc}
     &  & \multicolumn{4}{c|}{Qwen2.5-7B-base} & \multicolumn{4}{c}{Llama3.2-3B-Instruct} \\
    Config & Adapt. & IFEval & NQ & BBH & Avg. & IFEval & NQ & BBH & Avg. \\
    \shline
    SFT/Base          & Math & 45.32 & 26.40 & 69.18 & 46.97 & 77.58 & 31.75 & 49.75 & 53.03 \\
    SFT/Base          & Med  & 45.32 & 26.40 & 69.18 & 46.97 & 79.26 & 31.75 & 59.28 & 56.76 \\
    SFT/Full FT       & Math & 36.69 &  2.13 & 63.85 & 34.22 & 66.79 & 30.17 & 22.52 & 39.83 \\
    SFT/Full FT       & Med  & 36.33 &  0.86 & 66.04 & 34.41 & 46.40 & 12.80 & 18.88 & 26.03 \\
    SFT/OFT-b16       & Math & 41.25 & 18.17 & 68.34 & 42.58 & 67.75 & 26.59 & 28.91 & 41.08 \\
    SFT/OFT-b16       & Med  & 41.25 & 25.43 & 68.58 & 45.09 & 69.78 & 29.00 & 24.12 & 40.97 \\
    SFT/OFT-b32       & Math & 41.13 & 24.74 & 67.10 & 44.37 & 66.67 & 26.32 & 29.19 & 40.73 \\
    SFT/OFT-b32       & Med  & 40.53 & 18.01 & 68.65 & 42.40 & 69.30 & 28.48 & 23.72 & 40.50 \\
    SFT/OFT-b64       & Math & 38.97 &  2.80 & 66.15 & 35.97 & 66.55 & 26.76 & 25.93 & 39.75 \\
    SFT/OFT-b64       & Med  & 38.61 &  9.64 & 69.08 & 39.11 & 65.23 & 28.45 & 19.43 & 37.70 \\
    SFT/OFT-b128      & Math & 37.53 &  7.81 & 65.58 & 36.98 & 63.31 & 24.90 & 20.57 & 36.26 \\
    SFT/OFT-b128      & Med  & 37.77 &  3.60 & 69.28 & 36.88 & 56.47 & 27.31 & 18.98 & 34.26 \\
    SFT/LoRA-r4a8     & Math & 38.13 & 19.39 & 67.47 & 41.66 & 58.39 & 27.15 & 21.84 & 35.79 \\
    SFT/LoRA-r4a8     & Med  & 35.13 &  6.40 & 67.73 & 36.42 & 52.76 & 21.63 & 21.13 & 31.84 \\
    SFT/LoRA-r8a16    & Math & 34.53 & 16.70 & 66.43 & 39.22 & 63.07 & 26.62 & 20.02 & 36.57 \\
    SFT/LoRA-r8a16    & Med  & 35.01 &  5.90 & 68.57 & 36.06 & 48.32 & 16.59 & 19.06 & 27.99 \\
    SFT/LoRA-r16a32   & Math & 34.41 &  3.80 & 66.52 & 34.91 & 61.63 & 27.65 & 23.37 & 37.55 \\
    SFT/LoRA-r16a32   & Med  & 33.33 &  3.68 & 67.55 & 34.86 & 48.44 & 20.03 & 19.06 & 29.18 \\
    SFT/LoRA-r32a64   & Math & 35.01 & 13.80 & 65.81 & 38.21 & 62.35 & 26.79 & 22.46 & 37.20 \\
    SFT/LoRA-r32a64   & Med  & 35.25 &  3.68 & 67.57 & 35.50 & 50.12 & 20.80 & 21.14 & 30.69 \\
    SFT/AdaLoRA-r8a16 & Math & 32.61 &  6.65 & 66.98 & 35.41 & 56.35 & 25.79 & 21.45 & 34.53 \\
    SFT/AdaLoRA-r8a16 & Med  & 32.97 & 11.16 & 67.88 & 37.34 & 60.43 & 26.98 & 21.46 & 36.29 \\
    SFT/PiSSA-r8a16   & Math & 33.57 &  6.87 & 33.88 & 24.78 & 24.70 &  1.52 &  3.01 &  9.74 \\
    SFT/PiSSA-r8a16   & Med  & 30.10 &  4.24 & 19.82 & 18.05 & 19.18 &  3.85 & 15.73 & 12.92 \\
    SFT/MiLoRA-r8a16  & Math & 37.65 &  7.59 & 67.62 & 37.62 & 61.99 & 25.62 & 19.14 & 35.59 \\
    SFT/MiLoRA-r8a16  & Med  & 34.17 &  5.24 & 68.22 & 35.88 & 48.32 & 18.75 & 20.61 & 29.23 \\
    SFT/KeepLoRA-r8   & Math & 41.73 & 20.30 & 69.22 & 43.75 & 68.11 & 27.56 & 26.54 & 40.74 \\
    SFT/KeepLoRA-r8   & Med  & 45.32 & 26.68 & 69.27 & 47.09 & 67.63 & 28.98 & 21.95 & 39.52 \\
    SFT/MiSS-r8       & Math & 39.45 & 14.60 & 63.30 & 39.12 & 58.75 & 20.36 & 22.68 & 33.93 \\
    SFT/MiSS-r8       & Med  & 36.45 &  3.96 & 62.87 & 34.43 & 49.04 & 23.60 & 22.48 & 31.71 \\
    SFT/MiSS-r64      & Math & 35.85 &  2.74 & 59.72 & 32.77 & 60.91 & 20.69 & 23.28 & 34.96 \\
    SFT/MiSS-r64      & Med  & 35.37 &  0.69 & 62.08 & 32.72 & 39.81 &  8.37 & 20.17 & 22.78 \\
    SFT/VeRA-r256     & Math & 46.04 & 27.12 & 68.59 & 47.25 & 72.66 & 31.80 & 34.85 & 46.79 \\
    SFT/VeRA-r256     & Med  & 45.92 & 25.79 & 69.31 & 47.01 & 72.18 & 31.66 & 42.98 & 48.94 \\
    SFT/DoRA-r8a16    & Math & 36.81 & 17.56 & 66.36 & 40.25 & 60.55 & 26.70 & 19.70 & 35.65 \\
    SFT/DoRA-r8a16    & Med  & 34.05 &  6.54 & 67.58 & 36.06 & 47.00 & 17.87 & 17.73 & 27.53 \\
    SFT/IA3           & Math & 43.17 & 23.49 & 67.46 & 44.71 & 70.14 & 29.53 & 37.49 & 45.72 \\
    SFT/IA3           & Med  & 48.44 & 27.15 & 69.16 & 48.25 & 72.30 & 29.97 & 34.73 & 45.67 \\
    RLVR/Full FT      & Math & 50.00 & 25.84 & 70.20 & 48.68 & 76.50 & 31.02 & 49.08 & 52.20 \\
    RLVR/Full FT      & Med  & 42.69 & 17.45 & 69.51 & 43.22 & 76.98 & 31.02 & 47.44 & 51.81 \\
    RLVR/OFT-b32      & Math & 49.64 & 27.51 & 69.55 & 48.90 & 74.94 & 28.50 & 46.68 & 50.04 \\
    RLVR/OFT-b32      & Med  & 46.40 & 26.48 & 68.84 & 47.24 & 76.50 & 30.58 & 49.85 & 52.31 \\
    RLVR/LoRA-r8a16   & Math & 48.44 & 26.95 & 69.40 & 48.27 & 77.10 & 31.36 & 48.06 & 52.17 \\
    RLVR/LoRA-r8a16   & Med  & 48.44 & 11.08 & 68.88 & 42.80 & 76.98 & 31.72 & 51.91 & 53.53 \\
  \end{tabular}
  \end{adjustbox}
  \caption{Detailed general-benchmark scores for the checkpoints in \autoref{tab:main-results}.}
  \label{tab:app:detailed-general-scores}
\end{table*}

\clearpage

\section{Additional Weight-Space Geometry Results}
\label{app:weight-geometry}

\subsection{Full Spectral Profiles and Retention Smoothness}
\label{app:full-spectral-profiles}
\label{app:retention-smoothness-validation}

To complement the qualitative fluctuation description in \autoref{sec:spectral_analysis}, we quantify how the retention-side fluctuation score correlates with actual retention performance across the main-results checkpoints. For a profile $s(i)$ and its local moving average $\bar{s}(i)$, we define
\begin{equation}
  \mathrm{Fluc}(s)=\frac{1}{n}\sum_{i=1}^{n}\left|s(i)-\bar{s}(i)\right|.
  \label{eq:spectral-fluctuation}
\end{equation}
The retention-side fluctuation score is the mean absolute deviation of the diagonal-projection retention spectrum from this local moving average; larger values indicate a spikier, less smooth retention profile. In our implementation, we use a moving-average window of size $w=5$ (equivalently, radius $r=2$), apply reflect padding at the two boundaries before computing the 1D moving average, and set the score to zero when the spectrum length is shorter than the window. \autoref{tab:app:retention-fluctuation-corr} shows that larger retention fluctuation is significantly associated with lower general performance and more forgetting. The correlation is computed on the full set of matched checkpoints and the SFT subset. Additional visualization of weight-space profiles in \autoref{fig:spectrum_analysis_add} shows that different PEFT parameterizations perturb the pretrained spectrum in qualitatively different ways. Full finetuning and PiSSA produce the most visible retention-side disruption, while OFT keeps a more coherent diagonal-projection profile. LoRA, MiSS, PiSSA, and AdaLoRA all show uneven update-energy allocation, matching the large adaptation-side fluctuation scores shown in the main-text fluctuation-score bar plot in \autoref{fig:spectral_analysis}. PiSSA is especially distinctive: its update is strongest on the principal components, but it also remains spiky and non-negligible across the remaining ranks rather than being confined to only the top singular directions. In contrast, OFT's profile is more structured under its orthogonal parameterization, and the RL variants generally show smaller and more coherent perturbations than their SFT counterparts.

\vspace{3mm}
\begin{table}[htbp]
  \centering
  \footnotesize
  \setlength{\abovecaptionskip}{4pt}
  \setlength{\belowcaptionskip}{12pt}
  \setlength{\tabcolsep}{11pt}
  \renewcommand{\arraystretch}{1.25}
  \begin{adjustbox}{max width=\linewidth}
  \begin{tabular}{l|ccccc}
     &  & \multicolumn{2}{c}{General score} & \multicolumn{2}{c}{Forgetting} \\
    Subset   & $n$ & Spearman $\rho$ & $p$-value & Spearman $\rho$ & $p$-value \\
    \shline
    All checkpoints & 88 & -0.622 & $9.3\times 10^{-10}$ & 0.557 & $9.8\times 10^{-8}$ \\
    SFT only        & 76 & -0.413 & $5.2\times 10^{-4}$  & 0.329 & $6.6\times 10^{-3}$ \\
  \end{tabular}
  \end{adjustbox}
\caption{Spearman correlation between retention-side fluctuation and retention metrics. Larger fluctuation is associated with lower retained general ability and more forgetting.}
  \label{tab:app:retention-fluctuation-corr}
\end{table}

\begin{figure}[h!]
  \centering
  \setlength{\abovecaptionskip}{3pt}
  \setlength{\belowcaptionskip}{8pt}
  \vspace{1mm}
  \includegraphics[width=0.99\linewidth]{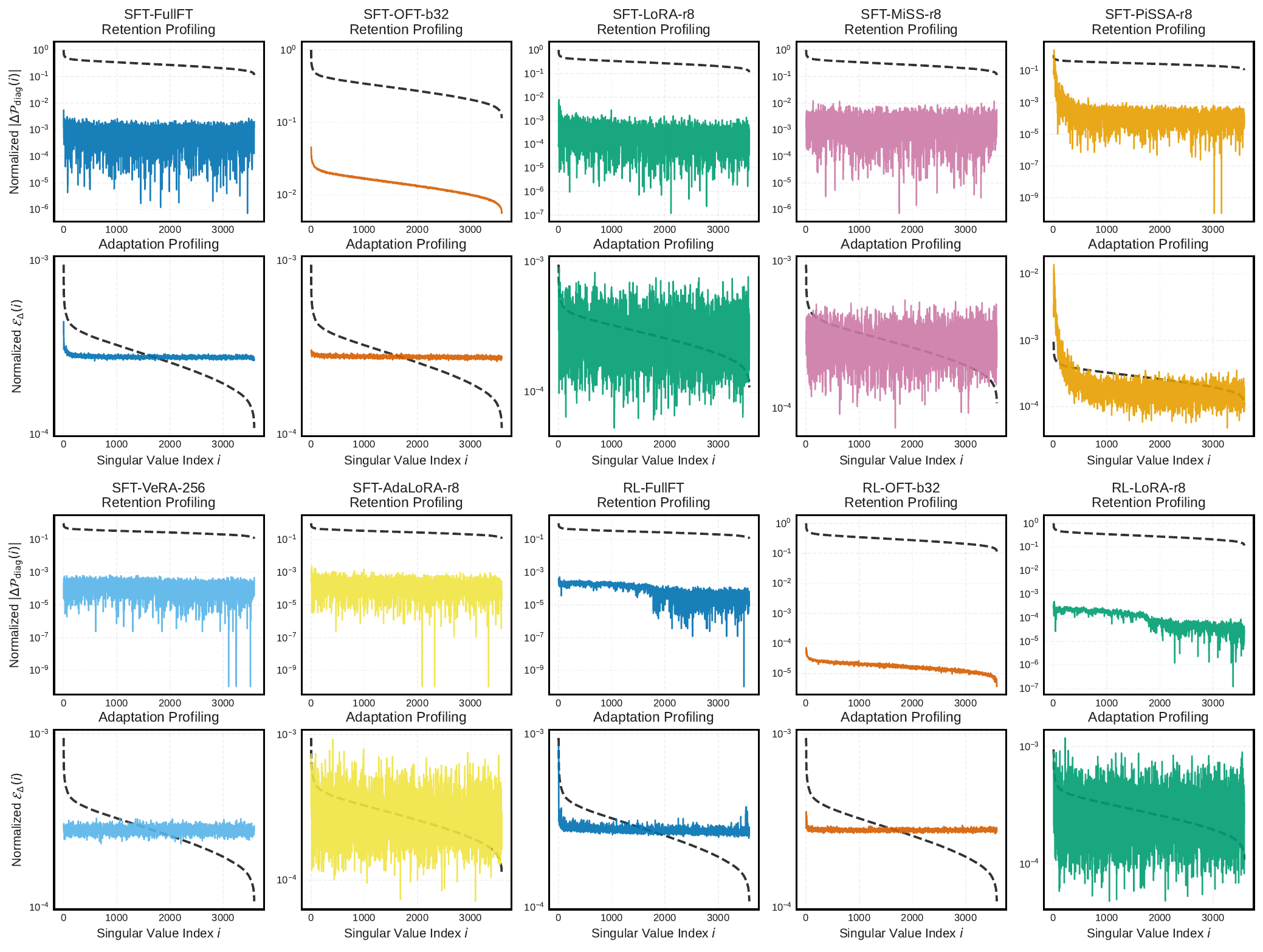}
  \caption{Additional weight-space profiles for SFT. We visualize diagonal-projection changes on the pretrained singular basis (retention profile) and projected update energy over pretrained directions (adaptation profile).}
  \label{fig:spectrum_analysis_add}
\end{figure}

\subsection{OFT Singular-Vector Alignment}
\label{app:oft-sva}

OFT requires a parameterization-aware interpretation because its effective update is an orthogonal transformation rather than a generic additive perturbation. The main text therefore focuses on diagnostics that can be compared across PEFT families: retention/adaptation profiles, capability-conditioned drift, and activation-geometry distortion. For completeness, we include an OFT-specific singular-vector alignment (SVA) diagnostic here to isolate the rotational component of OFT.

\begin{figure}[h!]
  \centering
  \setlength{\abovecaptionskip}{3pt}
  \setlength{\belowcaptionskip}{10pt}
  \vspace{3mm}
  \includegraphics[width=\linewidth]{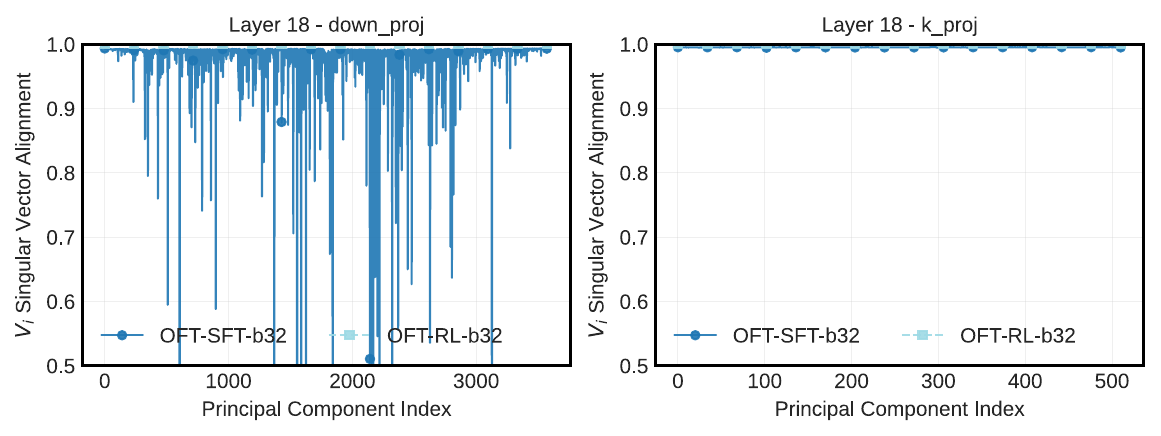}
  \caption{Additional OFT geometry diagnostic using singular vector alignment (SVA). OFT primarily changes singular-vector orientation while preserving singular values.}
  \label{fig:app-sva}
\end{figure}

Let $W_{0}=U_{0}\Sigma_{0}V_{0}^{\top}$. Under an ideal right-multiplicative OFT rotation,
\begin{equation*}
  W^{*}= W_{0}R=U_{0}\Sigma_{0}V_{0}^{\top}R,\quad R^{\top}R=I.
\end{equation*}
The singular values are preserved, while the right singular vectors rotate. We therefore compare the cosine similarity between corresponding pretrained and rotated singular vectors to measure how strongly OFT changes singular-vector orientation. As shown in \autoref{fig:app-sva}, RL-trained OFT exhibits a relatively uniform rotation pattern across components, consistent with a more global and coherent transformation of the layer weight. In contrast, SFT-trained OFT shows localized SVA spikes in specific layers and components, indicating that some singular directions receive disproportionately large rotations. This explains why OFT can mitigate forgetting through its orthogonal constraint while still showing residual retention loss under SFT: the update preserves singular values, but non-uniform rotations can still perturb directions used by general capabilities.

\subsection{Capability-Conditioned Drift}
\label{app:csd}

The retention and adaptation profiles in \autoref{sec:weight-geometry} describe an update in the pretrained singular basis, but they are not conditioned on which directions are actually used by a capability distribution. Capability-conditioned spectral drift (CSD) adds this data dependence. For a selected module, we collect the pretrained-model input activations $h$ on a dataset $D$, then measure the output displacement induced by the effective weight update $\Delta W=W^{*}-W_0$.

In our implementation, the selected module is layer 18 \texttt{mlp.down\_proj.weight}. For each checkpoint, $W^{*}$ is the effective finetuned weight: full-finetuned checkpoints are read directly, while PEFT adapters are merged into the base weight before extracting $W^{*}$. We use 161 general examples from IFEval/NQ/BBH for $D_G$, and 165--170 examples from the corresponding math or medical target domain for $D_T$. Activations are always collected from the pretrained model so that all methods are compared on the same capability-conditioned input distribution. The tables below report the SFT-only subset of the available CSD results.

For a dataset $D$, absolute CSD is defined as
\begin{equation*}
  \mathrm{CSD}_{\mathrm{abs}}(D)=
  \mathbb{E}_{h\sim D}\left[\lVert \Delta W h\rVert_2^2\right].
\end{equation*}
This quantity is a capability-conditioned projection of the update profile: if $h=\sum_i a_i v_i$ in the pretrained right singular basis, then $\Delta W h=\sum_i a_i\Delta W v_i$. Thus, directions that are both strongly updated and frequently activated by $D$ contribute more to CSD. We also report a relative version that controls for the scale of the pretrained module output,
\begin{equation*}
  \mathrm{CSD}_{\mathrm{rel}}(D)=
  \frac{\mathbb{E}_{h\sim D}\left[\lVert \Delta W h\rVert_2^2\right]}
  {\mathbb{E}_{h\sim D}\left[\lVert W_0 h\rVert_2^2\right]+\epsilon},
\end{equation*}
and an update-normalized version that partially controls for raw update magnitude,
\begin{equation*}
  \mathrm{CSD}_{\mathrm{un}}(D)=
  \frac{\mathbb{E}_{h\sim D}\left[\lVert \Delta W h\rVert_2^2\right]}
  {(\lVert \Delta W\rVert_F^2+\epsilon)
   (\mathbb{E}_{h\sim D}\left[\lVert h\rVert_2^2\right]+\epsilon)}.
\end{equation*}
We compute these quantities for the general distribution $D_G$ and the target distribution $D_T$, and summarize both $\mathrm{CSD}_G$ and the general-to-target ratio $\mathrm{CSD}_G/(\mathrm{CSD}_T+\epsilon)$. The ratio is intended to describe whether a given update perturbs general-capability activations disproportionately relative to target-domain activations. As with the spectral profiles, CSD is an empirical diagnostic rather than a causal proof.

\begin{table}[htbp]
  \centering
  \footnotesize
  \setlength{\abovecaptionskip}{4pt}
  \setlength{\belowcaptionskip}{12pt}
  \setlength{\tabcolsep}{7pt}
  \renewcommand{\arraystretch}{1.18}
  \vspace{3mm}
  \begin{tabular}{lccc}
    \textbf{Metric} & \textbf{External metric} & \textbf{Pearson} & \textbf{Spearman} \\
    \shline
    $\mathrm{CSD}_{G,\mathrm{rel}}$ & Forgetting  & 0.347 & 0.317 \\
    $\mathrm{CSD}_{G,\mathrm{un}}$ & Forgetting  & 0.481 & 0.603 \\
    $\mathrm{CSD}_{T,\mathrm{rel}}$ & Target gain  & -0.164 & 0.107 \\
    $\mathrm{CSD}_{T,\mathrm{un}}$ & Target gain  & -0.358 & -0.615 \\
    Ratio$_{\mathrm{rel}}$ & Forgetting  & -0.188 & -0.241 \\
  \end{tabular}
  \caption{CSD correlations over the SFT-only subset ($n=68$ checkpoints) at layer18 \texttt{mlp.down\_proj.weight}. General CSD is positively associated with forgetting, while target CSD does not provide a monotonic explanation of target gain.}
  \label{tab:app:csd-corr}
\end{table}

\begin{table}[htbp]
  \centering
  \footnotesize
  \setlength{\abovecaptionskip}{4pt}
  \setlength{\belowcaptionskip}{10pt}
  \setlength{\tabcolsep}{7pt}
  \renewcommand{\arraystretch}{1.18}
  \vspace{3mm}
  \begin{adjustbox}{max width=\linewidth}
  \begin{tabular}{lccccc}
    \textbf{Family} & \textbf{$n$} & $\mathbf{CSD}_{G,\mathrm{rel}}$ & $\mathbf{CSD}_{T,\mathrm{rel}}$ & \textbf{Ratio} & \textbf{Forget} \\
    \shline
    Full FT & 4  & $1.16{\times}10^{-3}$ & $1.06{\times}10^{-3}$ & 1.077 & 17.31 \\
    LoRA    & 16 & $1.21{\times}10^{-3}$ & $1.21{\times}10^{-3}$ & 1.042 & 15.70 \\
    OFT     & 16 & $7.82{\times}10^{-3}$ & $7.64{\times}10^{-3}$ & 1.018 & 11.20 \\
    PiSSA   & 4  & $8.94{\times}10^{-2}$ & $8.37{\times}10^{-2}$ & 1.102 & 34.56 \\
    MiLoRA  & 4  & $7.17{\times}10^{-4}$ & $6.46{\times}10^{-4}$ & 1.288 & 16.35 \\
    MiSS    & 8  & $1.93{\times}10^{-2}$ & $1.88{\times}10^{-2}$ & 1.019 & 18.13 \\
  \end{tabular}
  \end{adjustbox}
  \caption{Method-family CSD summary on the SFT-only subset. OFT has larger raw relative CSD than LoRA/Full FT because an orthogonal rotation can induce substantial Euclidean output movement, but its general-to-target ratio remains close to one and lower than Full FT, LoRA, PiSSA, and MiLoRA. PiSSA is the clearest outlier in both general CSD and forgetting.}
  \label{tab:app:csd-family}
\end{table}

\textbf{Interpretation.}
The CSD results complement the weight-space profiles in two ways. First, general-distribution CSD is a useful retention-side diagnostic: $\mathrm{CSD}_{G,\mathrm{rel}}$ is positively correlated with forgetting, and the update-normalized version remains positively correlated in rank order. This is consistent with the view that perturbing directions activated by general-evaluation data is associated with retention loss. Second, target-distribution CSD should not be used as a proxy for target gain; the correlation with target gain is weak or negative, consistent with the view that reasoning-heavy target improvements depend on task-aligned computation rather than update magnitude alone. Finally, raw CSD must be interpreted with the PEFT parameterization in mind. In particular, OFT can have higher Euclidean drift because orthogonal rotations move vectors, motivating the activation-geometry diagnostics used in the main text.

\section{Additional Activation-Space Geometry Results}
\label{app:activation-geometry}

\subsection{Metric Definitions}
\label{app:activation-metrics}

The main text focuses on Procrustes residual, CKA, and pairwise Gram distortion. Secondary diagnostics such as raw residual, angular drift, norm drift, and Procrustes improvement are used as controls to separate pointwise movement from non-isometric representation distortion.

For centered activations $X_0$ and $X_1$, Procrustes residual is
\begin{equation*}
  d_{\mathrm{proc}}=
  \frac{\min_{R^\top R=I}\|X_1R-X_0\|_F}{\|X_0\|_F+\epsilon}.
\end{equation*}
Linear CKA is computed from centered activations:
\begin{equation*}
  \mathrm{CKA}(X_0,X_1)=
  \frac{\|X_0^\top X_1\|_F^2}
  {\|X_0^\top X_0\|_F\,\|X_1^\top X_1\|_F+\epsilon}.
\end{equation*}
For row-normalized token activations $Z_0$ and $Z_1$, pairwise Gram distortion is
\begin{equation*}
  d_{\mathrm{gram}}=
  \frac{\|Z_1Z_1^\top-Z_0Z_0^\top\|_F}{\|Z_0Z_0^\top\|_F+\epsilon}.
\end{equation*}

\subsection{Coverage and Experimental Setup}
\label{app:activation-coverage}

We run the activation-geometry analysis on the SFT subset of the main-table checkpoints for Qwen2.5-7B and Llama3.2-3B-Instruct. For each checkpoint, we compare base and finetuned full-forward module outputs on general data (IFEval/NQ/BBH) and the corresponding target-domain data. We use eight module locations: layer 9 and layer 18, each with \texttt{q\_proj}, \texttt{k\_proj}, \texttt{v\_proj}, and \texttt{mlp.down\_proj}. In total, the compiled tables use 20 SFT checkpoints and 160 general-distribution rows for the correlation analysis.

\subsection{Layer/Module Method-Family Summaries}
\label{app:activation-layer-module}
\begin{table}[htbp]
  \centering
  \small 
  \setlength{\abovecaptionskip}{4pt}
  \setlength{\belowcaptionskip}{12pt}
  \setlength{\tabcolsep}{3.8pt}
  \renewcommand{\arraystretch}{1.2}
  \vspace{3mm}
  \begin{adjustbox}{max width=\linewidth}
  \begin{tabular}{llccccc}
    \textbf{Layer} & \textbf{Module} &
    \makecell{\textbf{Full FT}\\{Forgetting=17.31}\\{Proc.~/~Gram~/~CKA}} &
    \makecell{\textbf{LoRA}\\{Forgetting=15.97}\\{Proc.~/~Gram~/~CKA}} &
    \makecell{\textbf{OFT}\\{Forgetting=7.81}\\{Proc.~/~Gram~/~CKA}} &
    \makecell{\textbf{MiLoRA}\\{Forgetting=16.35}\\{Proc.~/~Gram~/~CKA}} &
    \makecell{\textbf{PiSSA}\\{Forgetting=34.56}\\{Proc.~/~Gram~/~CKA}} \\
    \shline
    9 & self\_attn.q\_proj &
    0.169~/~0.029~/~0.965 & 0.174~/~0.025~/~0.965 & \textbf{0.114~/~0.014~/~0.985} & 0.172~/~0.024~/~0.965 & 0.625~/~0.137~/~0.524 \\
    9 & self\_attn.k\_proj &
    0.182~/~0.040~/~0.971 & 0.182~/~0.037~/~0.974 & \textbf{0.118~/~0.027~/~0.990} & 0.175~/~0.036~/~0.978 & 0.665~/~0.217~/~0.609 \\
    9 & self\_attn.v\_proj &
    0.195~/~0.195~/~0.931 & 0.203~/~0.235~/~0.921 & \textbf{0.138~/~0.141~/~0.966} & 0.202~/~0.239~/~0.926 & 0.649~/~1.370~/~0.369 \\
    9 & mlp.down\_proj &
    0.117~/~0.211~/~0.898 & 0.121~/~0.219~/~0.902 & \textbf{0.074~/~0.151~/~0.964} & 0.117~/~0.219~/~0.915 & 0.452~/~1.598~/~0.518 \\
    18 & self\_attn.q\_proj &
    0.190~/~0.016~/~0.951 & 0.195~/~0.016~/~0.955 & \textbf{0.145~/~0.011~/~0.980} & 0.190~/~0.016~/~0.960 & 0.521~/~0.066~/~0.627 \\
    18 & self\_attn.k\_proj &
    0.201~/~0.049~/~0.966 & 0.205~/~0.042~/~0.970 & \textbf{0.160~/~0.026~/~0.983} & 0.204~/~0.037~/~0.974 & 0.570~/~0.169~/~0.697 \\
    18 & self\_attn.v\_proj &
    0.211~/~0.212~/~0.913 & 0.223~/~0.239~/~0.909 & \textbf{0.172~/~0.169~/~0.951} & 0.211~/~0.246~/~0.923 & 0.536~/~0.752~/~0.522 \\
    18 & mlp.down\_proj &
    0.164~/~0.250~/~0.865 & 0.181~/~0.243~/~0.856 & \textbf{0.128~/~0.191~/~0.934} & 0.164~/~0.248~/~0.865 & 0.438~/~0.865~/~0.440 \\
  \end{tabular}
  \end{adjustbox}
\caption{Activation-geometry method-family averages on general-evaluation data across tested layer/module locations for SFT checkpoints. Each cell reports Proc./Gram/CKA; lower Proc. and Gram and higher CKA indicate better geometry preservation. Column headers report method-family forgetting.}
  \label{tab:app:activation-full}
\end{table}

The full results of activation geometry method-family averages are shown in \autoref{tab:app:activation-full}. These metrics are primarily retention-side diagnostics. They do not show a simple monotonic relationship with target-domain gain, likely because reasoning-heavy target improvements depend on task-aligned computation and answer margins rather than representation movement magnitude alone.

\section{Interpolation and Rewinding Follow-ups}
\label{app:interpolation-followups}

\subsection{Interpolation Setup}
\label{app:interpolation-setup}

This appendix provides implementation details and additional results for the interpolation analyses in \autoref{sec:interpolation}.

\textbf{Additive-update PEFT (\eg, LoRA, AdaLoRA, MiSS)}. For PEFT methods with an additive parameterization,
\begin{equation*}
  W^{*}=W_0+\Delta W,\qquad W(\alpha)=W_0+\alpha\,\Delta W.
\end{equation*}
Interpolation is implemented by scaling the learned update $\Delta W$.
For LoRA, $\Delta W = s\,BA$ (with method-dependent scale $s$, \eg, $s=\tfrac{\alpha_{\mathrm{LoRA}}}{r}$). A simple implementation is to scale both factors,
\begin{equation*}
  \Delta W(\alpha) = s\,(\sqrt{\alpha}B)(\sqrt{\alpha}A)=\alpha\,s\,BA,
\end{equation*}
which preserves the product structure and avoids excessively scaling a single factor.

\textbf{OFT: interpolating rotation by scaling the generator}.
Using the Cayley parameterization, the rotation is hence parameterized by
\begin{equation*}
  R(Q)=(I+Q)(I-Q)^{-1},
\end{equation*}
where $Q$ is the skew-symmetric generator. We interpolate OFT by scaling its generator,
\begin{equation*}
  \begin{aligned}
  R(\alpha)&\triangleq R(\sqrt{\alpha} Q),\\
  R(\sqrt{\alpha} Q)&=(I+\sqrt{\alpha} Q)(I-\sqrt{\alpha} Q)^{-1}.
  \end{aligned}
\end{equation*}
Here $\alpha$ denotes the coefficient applied to rotation strength rather than directly to the generator norm. For small angles, the rotation angle $\theta$ satisfies $\|Q\|\approx \tan(\theta/2)\approx \theta/2$, hence $\theta$ is approximately linear in $\|Q\|$.
Following prior geometric diagnostics, we define the layer-wise rotation strength
\begin{equation*}
  \rho_{\ell}\triangleq 1-\cos(\theta_{\ell}).
\end{equation*}
Using $\cos\theta\approx 1-\theta^2/2$ gives $\rho_{\ell}\approx \theta_{\ell}^2/2$, and therefore $\rho_{\ell}\propto \|Q\|^2$.
Consequently, to scale the rotation strength by a factor of $\alpha$ in the small-angle regime, we scale the generator by $\sqrt{\alpha}$: $Q' = \sqrt{\alpha}\,Q$.

\newpage
\subsection{Full SFT Interpolation Curves}
\label{app:sft-interpolation}

\begin{figure}[h!]
  \centering
  \setlength{\abovecaptionskip}{3pt}
  \setlength{\belowcaptionskip}{5pt}
  \vspace{3mm}
  \includegraphics[width=0.99\linewidth]{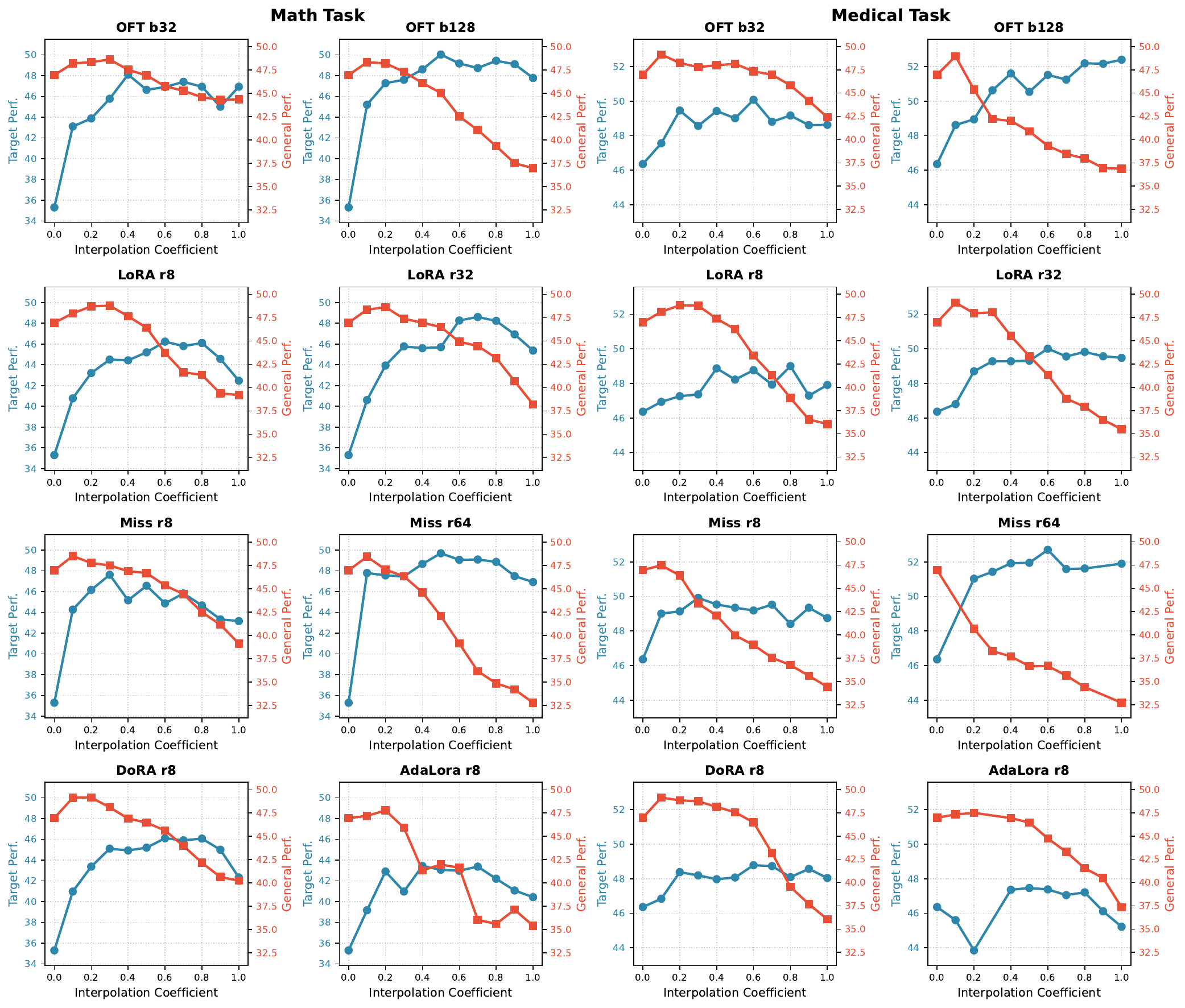}
  \captionof{figure}{Full SFT interpolation curves across PEFT methods and target domains. These results complement the representative pathwise interpolation analysis in the main text.}
  \label{fig:interpolate-extra}
\end{figure}

\autoref{fig:interpolate-extra} provides the full SFT interpolation curves across PEFT methods and target domains. These curves complement the single-column interpolation summary in the main text.

\subsection{OFT Linear vs Cayley Path}
\label{app:oft-cayley}

\autoref{tab:oft-q-vs-delta-app} reports the full OFT dense-delta versus Cayley-generator interpolation curve referenced in the main text.

\begin{table}[h!]
  \centering
  \footnotesize
  \setlength{\abovecaptionskip}{3pt}
  \setlength{\belowcaptionskip}{5pt}
  \setlength{\tabcolsep}{10pt}
  \renewcommand{\arraystretch}{1.2}
  \vspace{1mm}
\begin{tabular}{lccc}
\textbf{Path} & $\boldsymbol{\alpha}$ & \textbf{Math Avg.} $\uparrow$ & \textbf{General} $\uparrow$ \\
\shline
Delta interp. & 0.1 & 43.97 & 43.93 \\
Delta interp. & 0.3 & 43.93 & 43.91 \\
Delta interp. & 0.5 & 43.93 & 44.45 \\
Delta interp. & 0.7 & 44.60 & 43.61 \\
Delta interp. & 0.9 & 44.17 & 43.97 \\
\shline
$\sqrt{\alpha}Q$ interp. & 0.1 & 43.10 & 48.21 \\
$\sqrt{\alpha}Q$ interp. & 0.3 & 45.77 & \textbf{48.64} \\
$\sqrt{\alpha}Q$ interp. & 0.5 & 46.63 & 46.93 \\
$\sqrt{\alpha}Q$ interp. & 0.7 & \textbf{47.40} & 45.28 \\
$\sqrt{\alpha}Q$ interp. & 0.9 & 45.00 & 44.32 \\
Final OFT & 1.0 & 46.93 & 44.37 \\
\end{tabular}
  \caption{OFT interpolation path comparison on Qwen2.5-7B math SFT with OFT-b32. Dense-weight interpolation scales $W^{*}-W_0$, while the OFT path rescales the Cayley generator as $\sqrt{\alpha}Q$.}
  \label{tab:oft-q-vs-delta-app}
\end{table}

\newpage
\subsection{Layer-wise Rewinding Results}
\label{app:layerwise-rewinding}

\begin{figure}[h!]
  \centering
  \setlength{\abovecaptionskip}{3pt}
  \setlength{\belowcaptionskip}{10pt}
  \vspace{3mm}
  \includegraphics[width=0.99\linewidth]{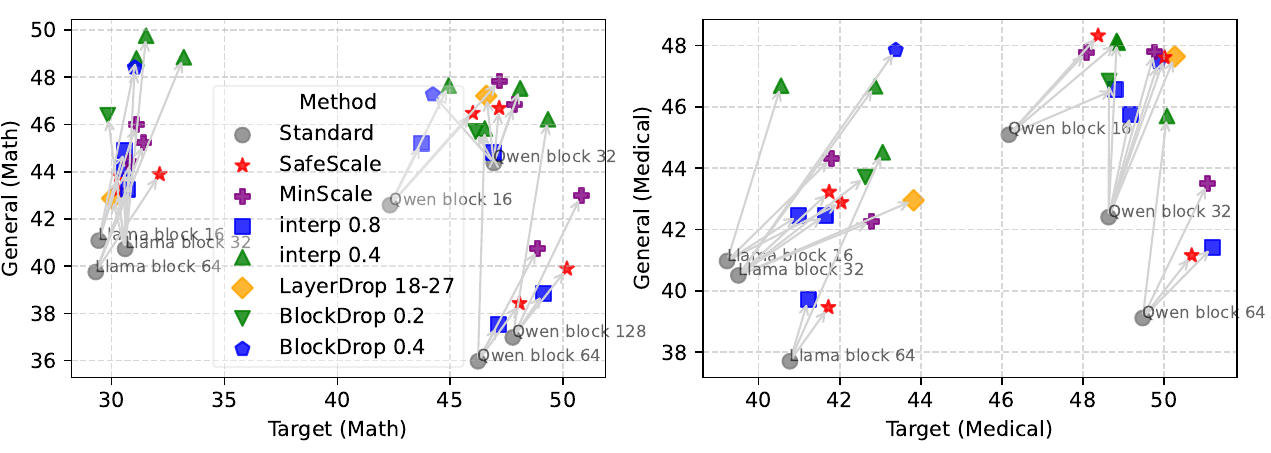}
  \caption{Trade-off alternatives achieve consistent improvement on OFT adapters of different sizes. This figure complements \autoref{tab:ablation-trade-off} by visualizing how different scaling/pruning alternatives behave across adapter sizes.}
  \label{fig:rewinding-tradeoff}
\end{figure}

In the main text, \autoref{tab:ablation-trade-off} reports the OFT layer-wise rewinding case study. \autoref{fig:train-vs-interp} compares actual optimization trajectories with interpolation trajectories, while \autoref{fig:rewinding-tradeoff} visualizes these OFT rewinding alternatives.

\begin{figure}[h!]
  \centering
  \vspace{3mm}
  \setlength{\abovecaptionskip}{3pt}
  \setlength{\belowcaptionskip}{10pt}
  \includegraphics[width=0.62\linewidth]{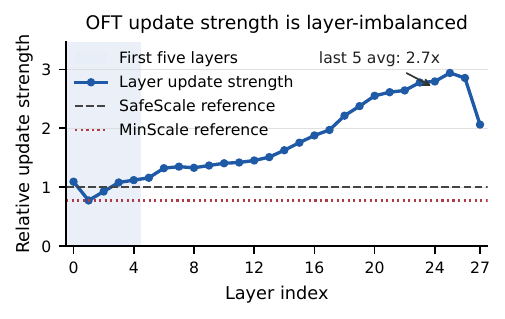}
  \caption{Layer-wise OFT update strength for Qwen2.5-7B math SFT with OFT-b32. Values are normalized by the average strength of the first five layers, which is the SafeScale reference.}
  \label{fig:oft-layer-update-strength}
\end{figure}

\autoref{fig:oft-layer-update-strength} shows the layer-wise OFT update strength used by SafeScale and MinScale. The strength is computed with the same statistic as the rewinding implementation: the average squared Frobenius norm of OFT generator parameters within each layer. Later layers receive substantially larger generator updates than the early layers, motivating layer-wise rewinding rather than a single global coefficient.

\newpage
\subsection{Additive PEFT Rewinding Variants}
\label{app:additive-rewinding}

\autoref{tab:additive_layerwise_followup_app} shows that beyond OFT, similar layer-wise rewinding variants can also improve additive PEFT methods in this setting.

\begin{table}[h!]
  \centering
  \footnotesize
  \setlength{\abovecaptionskip}{4pt}
  \setlength{\belowcaptionskip}{10pt}
  \setlength{\tabcolsep}{9pt}
  \renewcommand{\arraystretch}{1.2}
  \vspace{3mm}
\begin{tabular}{llcc}
\textbf{Method} & \textbf{Variant} & \textbf{Math Avg.} $\uparrow$ & \textbf{General} $\uparrow$ \\
\shline
LoRA-r8 & Baseline & 42.47 & 39.22 \\
LoRA-r8 & SafeScale & 43.87 & 39.55 \\
LoRA-r8 & MinScale & \textbf{44.93} & \textbf{43.11} \\
MiSS-r8 & Baseline & 43.17 & 39.12 \\
MiSS-r8 & SafeScale & 42.60 & 38.95 \\
MiSS-r8 & MinScale & \textbf{44.83} & \textbf{42.85} \\
\end{tabular}
  \caption{Additive PEFT rewinding variants on Qwen2.5-7B math SFT. MinScale improves both math and General for LoRA-r8 and MiSS-r8, suggesting that layer-wise rewinding is not limited to OFT.}
  \label{tab:additive_layerwise_followup_app}
\end{table}

\subsection{Longer-RLVR High-k Interpolation}
\label{app:rlvr-highk-interp}

\autoref{tab:rlvr_highk_followup_app} summarizes high-$k$ interpolation follow-ups, where interpolation exposes a clearer overshoot phenomenon than the standard pass@1 view.

\begin{table}[h!]
  \centering
  \footnotesize
  \setlength{\abovecaptionskip}{4pt}
  \setlength{\belowcaptionskip}{10pt}
  \setlength{\tabcolsep}{9pt}
  \renewcommand{\arraystretch}{1.2}
  \vspace{3mm}
\begin{tabular}{llcc}
\textbf{Setting} & \textbf{Variant} & \textbf{Pass@64} $\uparrow$ & \textbf{General} $\uparrow$ \\
\shline
OFT-200 & Baseline ($\alpha=1.0$) & 77.23 & 47.58 \\
OFT-200 & Best interp. ($\alpha=0.7$) & \textbf{78.07} & \textbf{48.39} \\
OFT-200 & MinScale & 76.13 & 47.37 \\
OFT-500 & Baseline ($\alpha=1.0$) & 75.77 & 46.93 \\
OFT-500 & Best interp. ($\alpha=0.7$) & \textbf{78.50} & \textbf{48.04} \\
OFT-500 & MinScale & 78.47 & 47.34 \\
LoRA-200 & Baseline ($\alpha=1.0$) & 76.73 & 44.64 \\
LoRA-200 & Best interp. ($\alpha=0.7$) & \textbf{77.43} & \textbf{47.31} \\
LoRA-500 & Baseline ($\alpha=1.0$) & 73.80 & 43.26 \\
LoRA-500 & Best interp. ($\alpha=0.3$) & \textbf{75.30} & \textbf{48.05} \\
\end{tabular}
  \caption{High-$k$ RLVR follow-ups under pass@64. Interpolation can improve both pass@64 and General in several longer-RLVR settings, exposing a pathwise degradation pattern not visible from pass@1 alone.}
  \label{tab:rlvr_highk_followup_app}
\end{table}

\end{document}